\documentclass[runningheads]{llncs}
\usepackage[T1]{fontenc}
\usepackage{graphicx}
\usepackage{booktabs}
\usepackage[misc]{ifsym}
\newcommand{\corr}{(\Letter)}
\usepackage{mwe}

\usepackage{comment}
\usepackage{amsmath,amssymb}
\usepackage{subcaption}

\usepackage{algorithm}
\usepackage{algpseudocode}
\usepackage{bm}
\usepackage{multirow}

\usepackage{xr}
\usepackage[hyphens]{url}

\usepackage{xstring}
\newcommand{\Eqref}{Eq.~\ref}
\newcommand{\citep}{\cite}
\newcommand{\citet}{\cite}
\newcommand{\qedsymbol}{\qed}
\newcommand{\aref}{Appendix~\ref}

\newtheorem{dfn}{Definition}
\newtheorem{prop}[dfn]{Proposition}

\makeatletter
\newcommand{\figcaption}[1]{\def\@captype{figure}\caption{#1}}
\newcommand{\tblcaption}[1]{\def\@captype{table}\caption{#1}}
\makeatother

\makeatletter
\let\OrigLabel\label
\let\OrigRef\ref
\let\OrigEqref\eqref
\let\OrigLtxlabel\ltx@label

\newcommand{\striphead}[2]{%
  \IfBeginWith{#1}{#2}{%
    \StrCut{#1}{#2}{\@unused}{\trimmedlabel}%
  }{%
    \def\trimmedlabel{#1}%
  }%
}

\newcommand{\PopLabel}[1]{%
  \let\PrevLabel\label
  \let\PrevLtxlabel\ltx@label
  \let\PrevRef\ref
  \def\ref##1{%
    \striphead{##1}{#1}%
    \PrevRef{\trimmedlabel}%
  }%
}

\newcommand{\PushLabel}[1]{%
  \let\PrevLabel\label
  \let\PrevLtxlabel\ltx@label
  \let\PrevRef\ref
  \def\label##1{%
    \PrevLabel{#1##1}%
  }%
  \def\ltx@label##1{%
    \PrevLtxlabel{#1##1}%
  }%
  \def\ref##1{%
    \PrevRef{#1##1}%
  }%
}

\newcommand{\ResetLabel}{%
  \let\label\OrigLabel
  \let\ltx@label\OrigLtxlabel
  \let\ref\OrigRef
  \let\eqref\OrigEqref
}

\newcommand{\mref}[1]{\OrigRef{#1}}
\newcommand{\mEqref}{Eq.~\mref}
\makeatother

\makeatletter
\providecommand\input@path{}
\g@addto@macro\input@path{{Supplementary/}}
\makeatother

\graphicspath{{./Supplementary/}}

\begin{document}

\title{TLXML: Task-Level Explanation of Meta-Learning via Influence Functions}
\titlerunning{Task-Level Explanation of Meta-Learning via Influence Functions}



\author{
    Yoshihiro Mitsuka\inst{1} \corr \and
    Shadan Golestan\inst{2} \and
    Zahin Sufiyan\inst{3} \and
    Shotaro Miwa\inst{1} \and
    Osmar R. Zaiane\inst{2, 3}
}


\authorrunning{Y. Mitsuka et al.}

\institute{
    Information Technology R\&D Center, Mitsubishi Electric Corporation
    \email{yoshihiro.mitsuka@gmail.com}
\and
    Alberta Machine Intelligence Institute
\and
    Department of Computing Science, University of Alberta
}

\tocauthor{
    Yoshihiro Mitsuka,
     Shadan Golestan,
     Zahin Sufiyan,
     Shotaro Miwa,
     Osmar R. Zaiane
}
\toctitle{TLXML: Task-Level Explanation of Meta-Learning via Influence Functions}

\maketitle
\let\oldthefootnote\thefootnote
\let\thefootnote\relax

\footnotetext{
This version of the contribution has been accepted for publication, after peer review (when applicable) but is not the Version of Record and does not reflect post-acceptance improvements, or any corrections.
The Version of Record is available online at: \url{http://dx.doi.org/10.1007/978-3-032-37664-0_22}.
Use of this Accepted Version is subject to the publisher’s Accepted Manuscript terms of use
\url{https://www.springernature.com/gp/open-research/policies/accepted-manuscript-terms}
.
}

\let\thefootnote\oldthefootnote
%
\begin{abstract}
Meta-learning enables models to rapidly adapt to new tasks by leveraging prior experience, but its adaptation mechanisms remain opaque, especially regarding how past training tasks influence future predictions. We introduce TLXML (Task-Level eXplanation of Meta-Learning), a novel framework that extends influence functions to meta-learning settings and provides task-level explanations of adaptation and inference. By reformulating influence functions for the bi-level structure of meta-learning, we quantify the contribution of each meta-training task to the adapted model’s behaviour. To ensure scalability, we propose a Gauss-Newton-based approximation that significantly reduces computational complexity from $O(pq^2)$ to $O(pq)$, where $p$ and $q$ denote the numbers of model and meta parameters, respectively. Moreover, we propose generalized influence functions defined using pseudo-inverse Hessian, which are applicable even when the loss landscape has flat directions. Results demonstrate that TLXML effectively ranks training tasks by their influence on downstream performance, offering concise, intuitive explanations aligned with user-level abstraction. This work provides a critical step toward interpretable and trustworthy meta-learning systems.
\end{abstract}
\section{Introduction}

Meta-learning, or "learning to learn", equips models with the ability to rapidly adapt to unseen tasks, addressing limitations in generalization caused by data scarcity during training~\citep{song2024towards,li2018learning,shu2021open}
and distribution shifts in deployment environments~\citep{mouli2024metaphysica,lin2020model}.
Despite its growing success and fast adaptation scenarios, meta-learning remains largely opaque. Current approaches offer limited insight into which training tasks influence the adaptation process and final predictions, creating significant barriers for transparency, trust,
 and safe deployment in real-world applications~\citep{khattar2023cmdp,wen2022improved}.
Consequently, robust explanation methods are required to enhance transparency and ensure safe operations of autonomous systems.

\begin{figure*}[t]
\centering
\includegraphics[width=\linewidth]{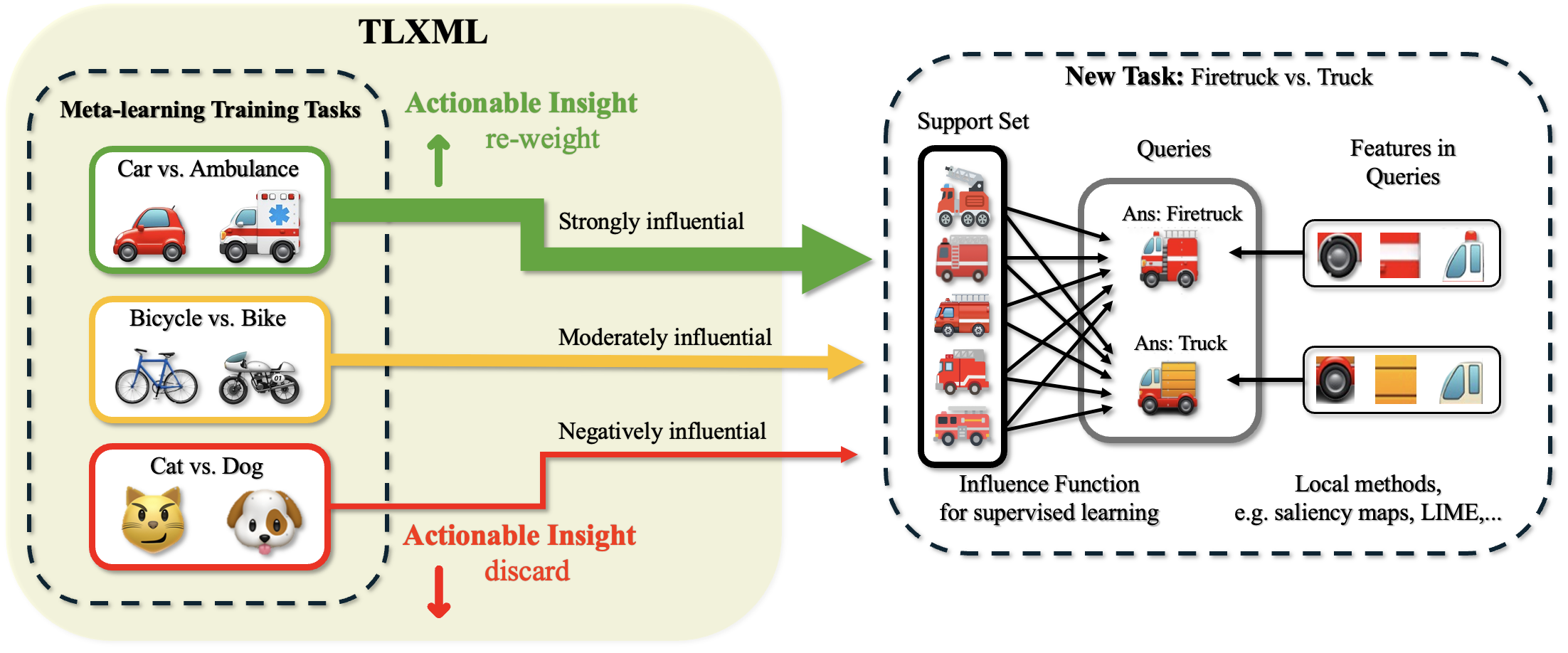}
\caption{Key insights of TLXML.}
\label{fig_intro}
\end{figure*}

Most explanation methods in machine learning focus on local interpretability, aiming to understand model behaviour around individual input data points. 
While such methods can be adopted to explain meta-learners' post-adaptation performances(the box on the right in Figure~\ref{fig_intro}), they fall short of capturing the unique characteristics of meta-learning. Notably, the final model behavior in meta-learning is highly sensitive not only to the test-time support set but also to the collection of prior training tasks, as highlighted by~\citet{agarwal2021sensitivity}.
Furthermore, local explanations often require technical expertise to interpret~\citep{adebayo2022post},
and may not be helpful for end-users seeking actionable or intuitive insights. In contrast, task-level explanations based on past experiences — those that attribute predictions to previously encountered tasks—align more naturally with how meta-learning models are trained and how humans reason about prior experience
(Figure~\ref{fig_intro}).
Understanding how individual tasks contribute to model behavior is especially important as future meta-learning systems will increasingly operate over large, heterogeneous, multi-domain datasets. 
Task-level insights can improve both the interpretability and safety of such systems. 

Influence functions~\citep{hampel1974influence,cook1980characterizations,koh2017understanding}
provide a powerful tool for tracing model predictions back to training data, and have been successfully used in standard supervised learning to identify influential data points. The variety of applications includes crafting/detecting adversarial training examples~\citep{cohen2020detecting},
quantifying model uncertainty~\citep{alaa2020discriminative}, and analyzing large language models~\citep{grosse2023studying}. 
Although those are considered applicable for explaining the influence of a data point used in the adaptation to each task, provided that the adaptation algorithm satisfies the underlying assumptions in those works (the box on the right in Figure~\ref{fig_intro}), they have not yet been extended to the meta-learning process.
In this paper, we propose TLXML (Task-Level eXplanation of Meta-Learning)—a novel framework that applies influence functions to the meta-learning setting. TLXML quantifies the impact of individual meta-training tasks on the model’s adaptation and inference behavior. 
We reformulate influence functions to accommodate the bi-level structure inherent in meta-learning algorithms. 
Our approach allows users to understand how prior tasks shape the learning process in a principled and interpretable way.

\paragraph{Contributions.}
Our main contributions are as follows:
1) Task-Level Explanations for Meta-Learning: We introduce TLXML, a principled method for quantifying the influence of meta-training tasks on adaptation and prediction in meta-learning, offering concise, interpretable explanations aligned with user abstraction levels.
2) Scalable Influence Computation: We analyze the computational cost of TLXML, showing that the exact method scales poorly with
$O(pq^2)$, where $p$ and $q$ are the number of model and meta-parameters, and propose a Gauss-Newton-based approximation that reduces the cost to $O(pq)$ for practical use of TLXML.
In addition, we generalize influence functions by using the pseudo-inverse Hessian to handle flat directions around the loss-minimization point.
3) Empirical Validation: We empirically demonstrate that TLXML can meaningfully rank meta-training tasks by their influence on adaptation and inference, successfully identifying helpful versus unhelpful training tasks across several benchmarks,
and it can be used to improve the adaptation of meta-trained models.

\paragraph{Code for experiments.}
\url{https://github.com/ymitsuka/TLXML}

\section{Related work}
\paragraph{Influence functions for supervised learning.}
This work extends the influence functions proposed for supervised learning~\citep{koh2017understanding} to meta-learning settings.
As in supervised learning~\citep{khanna2019interpreting,Barshan2020relatif}, we will see Fisher information metric offers a useful geometric perspective on influence functions for meta-learning. We will also see that it plays a role peculiar to meta-learning in handling a computational cost challenge.
We also note that the non-convexity of loss functions is argued to lead to erroneous influence functions in supervised learning~\citep{basu2021influence}. We will see that the same holds in meta-learning and address it by generalizing the definition of the influence functions.

\paragraph{Explainable AI (XAI) for meta-learning.} 
Naturally, XAI methods that are agnostic to the learning process are applicable for explaining inference in meta-learning. 
Although this area is still in its early stages,
several research examples exist~\citep{woznica2021towards,sijben2024function,shao2023effect}.
With a motivation close to ours, \citet{woznica2021towards} quantified the importance of high-level characteristics of a dataset, such as the size, the number of features, and the number of classes.
However, we emphasize that the strength of meta-learning lies in its ability to generalize across \textit{tasks}, and understanding the impact of training tasks is crucial for evaluating a model's adaptability. 

\paragraph{Task-relatedness.}
Apart from the breadth of potential applications of TLXML, one restricted application is quantifying the relatedness between training and test tasks.
Task-relatedness is a frequently discussed subject in multi-task learning and meta-learning.
Known quantification methods include ones that use distance or divergence measures between data distributions in tasks~\citep{shui2019principled}, the loss gradient for each task~\citep{chen2018gradnorm,yu2020gradient}, transferability from one task to another~\citep{zamir2018taskonomy,phuoc2022task}, task-embeddings obtained by fine-tuning a model with each task~\citep{achille2019task2vec}, and annotations of additional task attributions~\citep{hu2023understanding}.
Compatibility with the motivation of this work requires that the quantification be achieved without access to the training tasks or additional annotation in user environments.
We will compare TLXML-based task similarity with Task2Vec~\citep{achille2019task2vec}, as a well-established method that satisfies these requirements.

\paragraph{Machine Unlearning (MU).}
Removing the influence of training data from a trained model represents a valuable application of influence functions beyond increasing the interpretability of machine learning models. 
Aiming to handle privacy concerns in supervised learning, ~\citet{cao2015towards} introduced MU as a method of forgetting a training point 
without incurring the cost of retraining the model from scratch, and subsequent research established close connections to influence functions and utilize them for MU ~\citep{guo2020certified,Golatkar2020eternal,Golatkar2021mixed}.
See \citet{li2025machine} for a recent review that includes those progresses.
We will see that TLXML extends MU to meta-learning and can be used to both block and amplify the influence of training tasks.

\paragraph{Extension of TLXML.}
After the core ideas of this work were presented in earlier preprints, a related work \citet{ren2025evaluating} appeared.
After revisiting the argument in this work on the definition of task-level influence function and its computational barriers, it suggests replacing the Hessian with a diagonal matrix constructed from the loss-gradient magnitude vector. Although its theoretical justification is awaited, the proposed proxy for the influence function has been empirically shown to be useful for identifying harmful tasks for MAML models.
Another notable aspect not covered by this work is the extension to the influence of individual data points on each task, motivated by the goal of improving meta-training. In contrast, the motivation for this work is to provide users with concise explanations aligned with user-level abstraction, prioritizing task-level explanations over instance-level explanations.

\paragraph{}
See \aref{supp-sup: related_work} for more details on related work.

\section{Preliminaries}
\paragraph{Influence functions.} 
~\citet{koh2017understanding} proposed influence functions for measuring the impact of a training data point on the outcomes of supervised learning, under the assumption that the trained weights $\hat \theta$ minimize the empirical risk:
\begin{align*}
\hat{\theta} & =
\mathop{\underset{\theta}{\rm{argmin}}} L\left(D^{{\rm train}},f_{\theta}\right)=
\mathop{\underset{\theta}{\rm{argmin}}}\frac{1}{n}\sum_{i=1}^{n}l\left(z_{i},f_\theta\right)
\end{align*}
where $f_\theta$ is the model to be trained, 
$D^{\rm{train}}{=}\{z_i\}_{i=1}^{n}$ is the training dataset 
consisting of $n$ pairs of an input $x_i$ and a label $y_i$,
i.e., $z_i{=}(x_i, y_i)$, 
and the total loss $L$ is the sum of the losses $l$ of each data point. 
The influence functions are defined with a perturbation $\epsilon$ for each data point $z_j$:
\begin{align*}
\hat{\theta}_{\epsilon, j}
&=
\mathop{\underset{\theta}{\rm{argmin}}} L_{\epsilon, j} \left(D^{{\rm train}},f_{\theta}\right) 
= 
\mathop{\underset{\theta}{\rm{argmin}}}
\frac{1}{n}\sum_{i=1}^{n} l\left(z_{i},f_\theta\right)+
\frac{\epsilon}{n} l\left(z_j,f_\theta\right),
\end{align*}
which is considered as a shift of the probability that $z_j$ occurs in the data distribution.
The influence of the data $z_j$ on the model parameter $\hat\theta$ is defined as its increase rate with respect to this perturbation:
\begin{align}
I^{{\rm param}} (j) & \overset{\rm def}{=}
\left.\frac{d\hat{\theta}_{\epsilon,j}}{d\epsilon}\right|_{\epsilon=0}
 = - \frac{1}{n}
 \left. H^{-1}\frac{\partial l\left(z_j,f_\theta\right)}{\partial\theta}\right|_{\theta=\hat{\theta}}
 \label{eq:inf_func_def}
\end{align}
where the Hessian is given by 
$H {=}\left. \partial_\theta \partial_\theta L \right|_{\theta=\hat{\theta}}$.
The influence on any differentiable function of $\theta$ is defined through the chain rule. 
For example, the influence 
on the loss of a test data $z_{\rm test}$
is calculated to be
\begin{align*}
I^{{\rm perf}} (z_{\rm test}, j)
&\overset{\rm def}{=}
\left.
\frac{dl\left(z_{\mathrm{test}}, f_{\hat{\theta}_{\epsilon,z_j}}\right)}{d\epsilon}
\right|_{\epsilon=0} 
=
\left.
\frac{dl\left(z_{\mathrm{test}},f_{\theta}\right)}{d\theta}
\right|_{\theta=\hat\theta}\cdot I^{{\rm param}}(j).
\end{align*}
An underlying assumption is that the Hessian matrix is invertible, 
which is not always true. 
Typically, over-parameterized networks have flat directions around loss-minimization points.
This issue is also addressed by this work.

\paragraph{Supervised meta-learning.}
In supervised meta-learning (See~\citet {Hospedales2022meta} for a review), 
a task $\mathcal{T}$ 
is defined as the pair of a dataset and a loss function: $\left(\mathcal{D}^{\mathcal{T}},\mathcal{L}^{\mathcal{T}}\right)$.
The occurrence of each task follows a distribution $\mathcal{T}{\sim}p\left(\mathcal{T}\right)$.
An adaptation algorithm $\mathcal{A}$ takes as input a task $\mathcal{T}$ 
and meta-parameters $\omega$, 
and outputs the adapted weights $\hat\theta$ of the model $f_{\theta}$.
The learning objective is stated as
the optimization of $\omega$ with respect to the loss averaged over the task distribution:
\begin{align*}
\hat{\omega}
& =\mathop{\underset{\omega}{\rm argmin}}\underset{\mathcal{T}\sim p(T)}{E}
 \left[
 \mathcal{L}^{\mathcal{T}}(
   \mathcal{D}^{\mathcal{T}},
   f_{\hat{\theta}^{\mathcal{T}}}
   )
 \right]\ \ \ \ 
\text{with}
\ \hat{\theta}^{\mathcal{T}}
=\mathcal{A}(\mathcal{T},\omega)
\end{align*}
The formulation of empirical risk minimization
uses sampled tasks as building blocks, which are divided into a taskset for training (source taskset) and a taskset for testing (target taskset):
\begin{align*}
D^{\rm src}{=}\{\mathcal{T}^{\rm{src}(i)}\}_{i=1}^{M},\ \ \ \ 
D^{\rm trg}{=}\{\mathcal{T}^{\rm{trg}(i)}\}_{i=1}^{M'}.
\end{align*}
The dataset in each task is also divided into a training and test dataset: $\mathcal{D}^{\mathcal{T}} = (\mathcal{D}^{\mathcal{T}\rm{train}}, \mathcal{D}^{\mathcal{T}\rm{test}})$.
The learning objective is stated as:
\begin{align}
\hat{\omega}
 & =\mathop{\underset{\omega}{\rm argmin}}\frac{1}{M}\sum_{i=1}^{M}\mathcal{L}^{\rm{src}(i)}(\mathcal{D}^{{\rm src}(i){\rm test}},f_{\hat{\theta}^{i}})\ \ \ \ 
 \text{with}\ \hat{\theta}^{i}=\mathcal{A}(\mathcal{D}^{{\rm src}(i){\rm train}},\mathcal{L}^{\rm{src}(i)}, \omega)
  \label{eq:meta_learning_gen}
\end{align}
An example performance metric in meta-testing used in this paper is the test loss
$\mathcal{L}^{\rm{trg}(i)}(\mathcal{D}^{{\rm trg}(i){\rm test}},f_{\hat{\theta}^{i}})$
where
$\hat{\theta}^{i}{=}\mathcal{A}(\mathcal{D}^{{\rm trg}(i){\rm train}}, \mathcal{L}^{\rm{trg}(i)}, \hat\omega)$.
We conduct experiments with MAML~\citep{finn2017model}, and Prototypical Network (Protonet)~\citep{snell2017prototypical}, two widely used meta-learning methods.
\aref{supp-sup: meta_learners} provides the explicit forms of $\mathcal{A}$ in those methods.

\section{Proposed Method} \label{method}
\subsection{Task-level Influence Functions}
We now describe our method.
TLXML measures the influence of training tasks 
on the adaptation and inference processes in meta-learning.
To measure the influence of a training task $\mathcal{T}^j$ on the model's behaviors, we consider the task-level perturbation of the empirical risk defined in \Eqref{eq:meta_learning_gen}:
\begin{align}
\hat{\omega}_{\epsilon}^{j}
=
\mathop{\underset{\omega}{\arg\min}}
&
\frac{1}{M}\sum_{i=1}^{M}
\mathcal{L}^{i}(\mathcal{D}^{(i){\rm test}},f_{\hat{\theta}^{i}}) 
+
\frac{\epsilon}{M}
\mathcal{L}^{j}(\mathcal{D}^{(j){\rm test}},f_{\hat{\theta}^{j}}) \label{eq:task_perturb}
\end{align}
with
$\hat{\theta}^{i}{=}\mathcal{A}(\mathcal{D}^{(i){\rm train}},\mathcal{L}^i, \omega)$.
The influence on $\hat\omega$ is defined as
\begin{align}
I^{{\rm meta}}(j)
&\overset{{\rm def}}{=} 
  \left.\frac{d\hat{\omega}_{\epsilon}^{j}}{d\epsilon}\right|_{\epsilon=0} 
=
-\frac{1}{M} \left.
H^{-1}
\frac{\partial\mathcal{L}^j(\mathcal{D}^{(j)\rm{test}},f_{\hat{\theta}^{j}})}{\partial\omega}
  \right|_{\omega=\hat{\omega}}
  \label{eq:influence_ml_gen_params}
\end{align}
where the Hessian matrix $H$ is defined as follows:
\begin{align}
H = \left.
    \frac{1}{M}\sum_{i=1}^{M}
      \frac{\partial^{2}\mathcal{L}^i(\mathcal{D}^{(i)\rm{test}},f_{\hat{\theta}^{i}})} 
           {\partial\omega\partial\omega}
    \right|_{\omega=\hat\omega}.
    \label{eq: hessian-ml-gen}
\end{align}
See~\aref{supp-sup: implicit_differentiation} 
for the derivation of 
\Eqref{eq:influence_ml_gen_params}.
The model's behavior is affected by 
the perturbation through the adapted parameters
$\hat\theta^{ij}_\epsilon \equiv \mathcal{A}(\mathcal{D}^{(i)\rm{train}}, \mathcal{L}^i, \hat\omega^j_\epsilon)$.
The influence of the training task $\mathcal{T}^j$ on model parameters 
$\hat\theta^{i}=\mathcal{A}(\mathcal{D}^{(i)\rm{train}}, \mathcal{L}^i, \hat\omega)$
is defined through the chain rule of differentiation as
\begin{align}
I^{{\rm adpt}}(i, j)
& \overset{{\rm def}}{=}
\left.\frac{d\hat{\theta}_{\epsilon}^{ij}}{d\epsilon}\right|_{\epsilon=0}
=
\left.
\frac{\partial\mathcal{A}
  \left(
  \mathcal{D}^{(i)\rm{train}},\mathcal{L}^i,\omega
  \right)}{\partial\omega}\right|_{\omega=\hat{\omega}}I^{{\rm meta}}(j),
\label{eq:influence_ml_gen_adpt}
\end{align}
and further the influence of the training task $\mathcal{T}^j$ on the loss of a test task $\mathcal{T}^i$ is defined as
\begin{align}
I^{{\rm perf}}(i, j) & \overset{{\rm def}}{=}
\left.
  \frac{d\mathcal{L}^i\left(
    \mathcal{D}^{(i)\rm{test}},f_{\hat{\theta}_{\epsilon}^{ij}}
    \right)}{d\epsilon}\right|_{\epsilon=0}
=
\left.\frac{\partial\mathcal{L}^i\left(\mathcal{D}^{(i)\rm{test}},f_{\theta}\right)}{\partial\theta}\right|_{\theta=\hat{\theta}{}^{(i)}}I^{{\rm adpt}}(i, j).
\label{eq:influence_ml_gen_perf}
\end{align}
The training tasks are used only for evaluating $I^{\rm meta}$. 
This implies we can obtain the other explanation data by retaining only the calculated result of $I^{\rm meta}$ without access to the raw training data, which helps us design devices with limited storage capacity.
We also note that the above method can optionally be extended beyond the task level when the task-level abstraction is insufficient.
See~\aref{supp-sup: task_grouping} for details. 

\subsection{Approximation via Gauss-Newton matrix}\label{hessian_approximation}
TLXML faces computational barriers when applied to large models due to the difficulties in handling a large Hessian $H$(\Eqref{eq: hessian-ml-gen}).
We note that, besides the computational cost of $\mathcal{O}(q^3)$ in inverting $H$ of matrix size $q$, there is a cost issue in computing $H$ itself, which is peculiar to meta-learning.
Although $H$ is defined as the second-order tensor of the meta parameters, the bi-level structure of meta-learning raises a third-order tensor
in the form $\partial_\omega \partial_\omega \theta$
during its computation, 
resulting in a computational cost of $\mathcal{O}(pq^2)$ 
for a model with $p$ weights and $q$ meta-parameters.
(see-\aref{supp-sup: explicit_forms} for details). 
To address the above issues, we extend the Gauss-Newton matrix (GN matrix) approximation method established in supervised learning (see, for example, \citep{martens2020new}) to meta-learning settings to handle the Hessian in the meta-parameter space.
For specific loss functions,
e.g., mean squared error and cross-entropy, the Hessian can be decomposed into 
a sum of outer products of two vectors 
along with terms containing second-order derivatives. 
In this work, we only work on the case of cross-entropy 
with the softmax function, which leads to:
\begin{align}
\frac{\partial^{2}L}{\partial\omega\partial\omega} 
 &=
 \ \ positive\ constant\times \nonumber\\
 &
 \left[
\sum_{njk}\sigma_{k}\left(\boldsymbol{y}_{n}\right)\left(\delta_{kj}-\sigma_{j}\left(\boldsymbol{y}_{n}\right)\right)\frac{\partial y_{nk}}{\partial\omega}\frac{\partial y_{nj}}{\partial\omega}
 -\sum_{njk}t_{nj}\left(\delta_{jk}-\sigma_{k}\left(\boldsymbol{y}_{n}\right)\right)\frac{\partial^{2}y_{nk}}{\partial\omega\partial\omega}
 \right]
\label{eq: preOPA}
\end{align}
where $y$ is the output of the last layer, 
$\sigma$ is the softmax function, 
$t$ is the one-hot vector of the target label, 
$j,k$ are class indices, 
and $n$ is the index specifying tasks and input tensors in them. 
The first term in \Eqref{eq: preOPA} is the GN matrix, or equivalently, the Fisher information metric. 
Since the second-order derivatives in the second term give rise to third-order tensors, we focus on cases where the second-order derivatives are uncorrelated with their coefficients and the Hessian is approximated by the first term.
In supervised learning, the loss $L$ and the model output $\mathbf{y}$ are functions of the model's weights $\theta$; in our setting, they depend on the meta-parameters $\omega$ through the adaptation process.
Based on basic facts regarding cross-entropy (see~\aref{supp-sup: OPA_proof}), we can reuse the argument in supervised learning and show that when the training taskset closely approximates
a distribution $P\left(X|\omega^{*}\right)$ with the optimal value $\omega^{*}$ and the learned $\hat\omega$ is close to $\omega^*$, the first term in \Eqref{eq: preOPA} dominates.
As the quadratic form of the gradient vectors for each $n$ in the first term is non-negative, we express it with the factorized form:
\begin{align}
\frac{\partial^{2}L}{\partial\omega_\mu\partial\omega_\nu}
&\sim
\sum_{nj}\left(\mathbf{V}\right)_{\mu(nj)}\left(\mathbf{V}^{{\rm T}}\right)_{(nj)\nu}
\label{eq: OPA}
\end{align}
where $\mu$ and $(nj)$ are regarded as a row index and a column index, respectively. In the case of $q$ meta-parameters, $M$ tasks, $N$ data points per task, and $c$ target classes, the shape of $\mathbf{V}$ is $q\times cNM$, meaning that, if $q > cNM$, the approximated Hessian has zero eigenvalues.  This also happens for smaller $q$ if the columns of $\mathbf{V}$ are not independent of each other.

To efficiently compute the inverse (or the pseudo-inverse defined below) of \Eqref{eq: OPA}, 
we orthogonalize the columns of $\mathbf{V}$. 
In \aref{supp-sup: implementation}, we show that, under a realistic assumption on the above-mentioned low-rank structure, this can be done sequentially by repeatedly orthogonalizing small sets of column vectors, avoiding the memory and computational costs of accumulating all terms and orthogonalizing a large set of vectors at once (\aref{supp-sup: implementation}, Proposition~\ref{supp-prop: OPA}).

\subsection{Influence Functions with Flat Directions} 
\begin{figure}[t]
  \centering
  \includegraphics[width=0.9\linewidth]{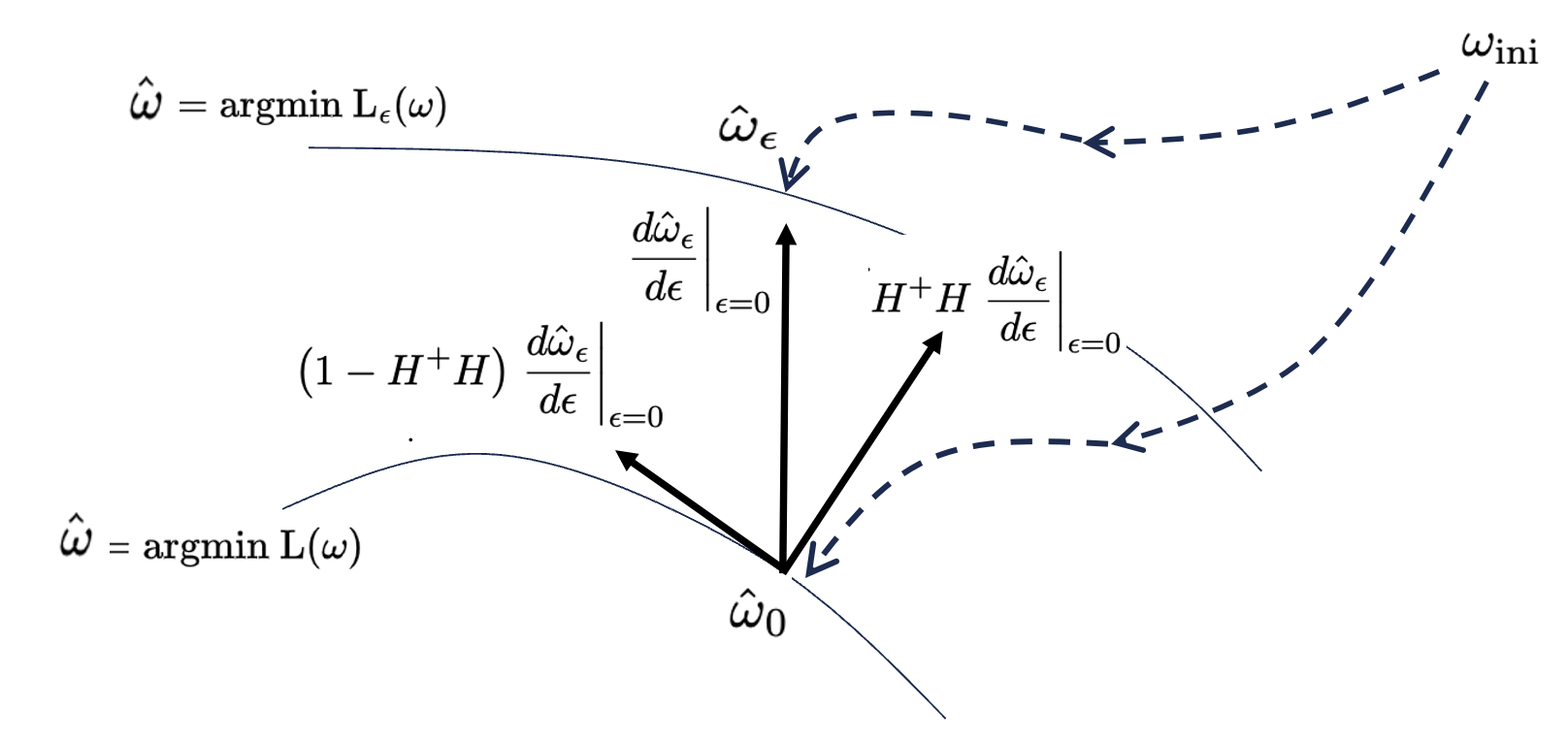}
  \caption{Diagram of the projected influence function, which measures the influence of a training task on the meta-parameters with the Hessian flat directions projected out.}
  \label{fig: pseudo_inverse}
\end{figure}

The appearance of a low-rank structure of the Hessian is not limited to the cases of the approximation mentioned above. Generally, it appears when the constraints imposed by the training tasks on the meta-parameters are insufficient to determine unique optimal values.
Figure~\ref{fig: pseudo_inverse} depicts a generic case where 
flat directions of the Hessian 
appear in the parameter space. 
When the number of parameters is sufficiently large,
the points that satisfy the minimization condition,
$\hat\omega=\rm{argmin}\ L(\omega)$,
form a hypersurface, resulting in flat directions of the Hessian.  
The same holds for the perturbed loss $L_\epsilon(\omega)$ 
used for defining the influence functions.
The position along the flat directions resulting from minimization depends on the initial conditions
and the learning algorithm.
We do not explore this dependency in our paper.
Instead, we employ a geometric definition of influence functions:  we take the partial inverse $H^+$ of $H$
in the subspace perpendicular to the flat directions,
known as the pseudo-inverse matrix.
With this approach, we modify the definitions of influence functions as:
\begin{align}
I^{{\rm meta}}(j)
& \overset{{\rm def}}{=}
  H^+ H\left.\frac{d\hat{\omega}_{\epsilon}^{j}}{d\epsilon}\right|_{\epsilon=0}
  = -\frac{1}{M}\left.
  H^{+}
\frac{\partial\mathcal{L}^j(D^{(j)\rm{test}},f_{\hat{\theta}^{j}})}{\partial\omega}
  \right|_{\omega=\hat{\omega}}.
  \label{eq:influence_ml_gen_param_mod}
\end{align}
See~\aref{supp-sup: implicit_differentiation} for the derivation of the second equation.
$H^+H$ represents the projection that drops the flat directions. 
When the Fisher information metric approximates the Hessian, the data distribution is unchanged in the flat directions, and the influence function projected by $H^+H$ reflects the sensitivity of $\hat \omega$ in the steepest direction of distribution change.
We also note that with that approximation (\Eqref{eq: OPA}), $H^+$ can also be written as a summation of factorized forms. 
This property is crucial for implementation
(\aref{supp-sup: implementation}, Proposition~\ref{supp-prop: pseudo_inv_OPA}).

In supervised learning, it is argued that the non-convexity of loss functions causes erroneous influence functions~\citep{basu2021influence}. The remedy of generalizing influence functions with the pseudo-inverse Hessian introduced here will also be effective in supervised learning.

\subsection{One-step Update via TLXML}
 Here, we consider utilizing TLXML to enhance the performance of meta-learners.
 In supervised learning,
 ~\citet{koh2017understanding}
 employed a leave-out approach, in which the network is retrained after removing training data with low influence scores.
 Applying this approach to improve meta-learning seems reasonable, but it digresses from the theoretical argument. 
 Since the influence functions are defined solely by the local structure around the convergence point, it is not guaranteed that tasks or data with low scores exert negative influences throughout the entire training process.
 
Recalling that $I^{{\rm meta}}$ represents the derivative of the meta-parameters with respect to the perturbation $\epsilon$ in the training task distribution, we can regard it as a linear approximation of the parameter shifts caused by that distribution change:
 \begin{align}
\delta \omega
&\sim 
  \xi \times \left.\frac{d\hat{\omega}_{\epsilon}^{j}}{d\epsilon}\right|_{\epsilon=0}
= 
\xi \times I^{{\rm meta}}(j)
\label{eq: one_step_update}
\end{align}
where we renamed the non-zero perturbation parameter to $\xi$ 
to avoid confusion with the differential variable. 
The case of $\xi{=}-1$ corresponds to the manipulation of removing task $j$ from the dataset, and if we are agnostic about the difference between supervised-learning and meta-learning, it is the situation considered by \citet{koh2017understanding,guo2020certified,Golatkar2020eternal,Golatkar2021mixed} for removing the influence of a specific training data point.
However, we should also note that $\xi$ can take other values as a parameter of the training task distribution.  \Eqref{eq: one_step_update} can be used not only to 
remove the influence of low-scored tasks but also to adjust it with other negative values of $\xi$, and even amplify the influence of high-scored tasks with positive $\xi$ without rerunning the training.

\section{Experiments}\label{experiments}
First, we examine whether TLXML produces explanation data that properly describes the influences of training tasks, and second, we examine whether the post-meta-training update using them improves adaptation ability.

\subsection{Validation of TLXML}
We investigate whether the proposed method 
satisfies two fundamental properties:
(Property 1)
If the network memorizes a training task, its influence on a test task with similar characteristics should be scored higher than the influence of other training tasks; and (Property 2) if the network encodes generalizable information about the task distribution, a training task subpopulation that shares salient features with the test task should be scored statistically higher than tasks outside that subpopulation.
Both properties are natural requirements for the method to be regarded as a method based on past experiences, as training tasks that resemble the test task
generally boost test performance.
We employ MAML and Protonet as meta-learning algorithms and conduct experiments on few-shot learning problems defined on the MiniImagenet~\citep{vinyals2016matching} and Omniglot~\citep {lake2015human} datasets, as well as two newly defined datasets (the synthetic dataset and FC 60, described below). 
Unless otherwise stated, experiments use a 5-way 5-shot configuration. 
The implementation builds on the \textit{lean2learn} meta-learning library \citep{arnold2020learn2learn} 
and PyTorch’s functions for automatic differentiation. 
We train the meta-parameters using Adam with a meta-batch size of 32.

\begin{figure}[t]
    \centering
    \begin{subfigure}[b]{0.49\textwidth}
        \centering
        \includegraphics[
        width=\textwidth]{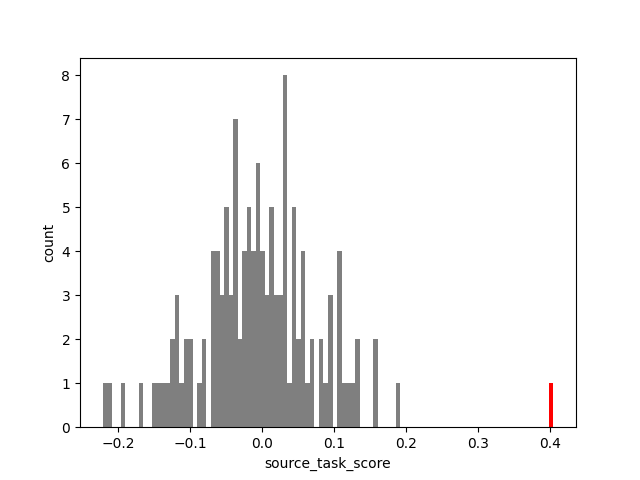}
        \caption{Influence scores for a single test.}
        \label{fig: test-with-training-tasks_a}
    \end{subfigure}
    \begin{subfigure}[b]{0.49\textwidth}
        \centering
        \includegraphics[
        width=\textwidth]{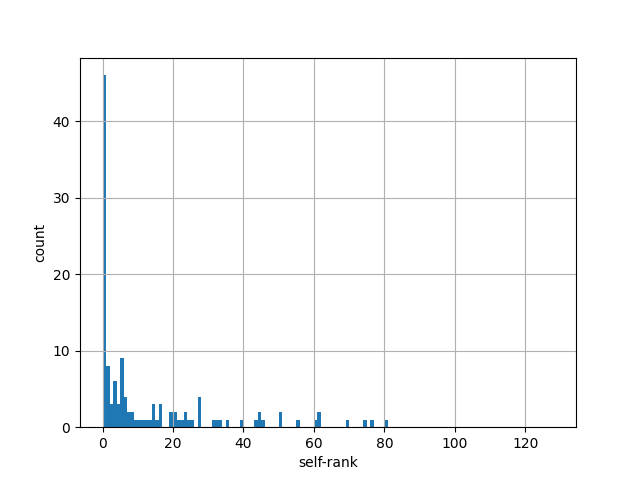}
        \caption{Self-ranks for 128 tests.}
        \label{fig: test-with-training-tasks_b}
    \end{subfigure}
    \caption{Test with training tasks. \textbf{Left: } Example of training task score distribution in a single test. Highlighted is the training task identical to the test task. \textbf{Right: } Histogram of the self-ranks of 128 test tasks
    }
    \label{fig: test-with-training-tasks}
\end{figure}

\subsubsection{Distinction of Tasks}

\label{task_distinction}
To validate Property 1, we generate imitations of pairs of similar training and test tasks by making each test task identical to a task in the training taskset. 
We then apply \Eqref{eq:influence_ml_gen_perf} to the trained network and examine whether it assigns the highest influence score to the training task identical to each test task.
Figure~\ref{fig: test-with-training-tasks} presents the results for a two-layer, fully connected network (1,285 parameters) overfitted to 128 MiniImageNet training tasks with MAML.
Figure~\ref{fig: test-with-training-tasks_a} illustrates a successful case of a test in which the identical training task (highlighted in red) is clearly distinguished from the other training tasks. 
Figure~\ref{fig: test-with-training-tasks_b} shows the distribution of the ranks assigned to the training tasks of this kind across 128 tests (we call them \textit{self-ranks}).
Although they often appear near the top of the ranking,
\footnote{In \aref{supp-sup: task_distinction}, as a sanity check, we see that reducing the similarity of training and test tasks leads to a degradation of the self-ranks.},
it is not always placed first(it is $12.6\pm18.9$).
This is likely because of the non-convexity of the training loss.
In our case, many of the 1285 Hessian eigenvalues 
are close to zero, and 92 are negative, violating the underlying assumption of \Eqref{eq:influence_ml_gen_params}.
The generalized influence function in~\Eqref{eq:influence_ml_gen_param_mod},
which replaces the inverse Hessian with its pseudo-inverse,
circumvents this instability.
Replacing the 92 negative eigenvalues with zero and inverting the only 1193 positive eigenvalues drives the self-rank to a perfect $0.0\pm0.0$.
The rank stays perfect as we aggressively truncate the spectrum down to 1024, 512, 256, 128, and 64.
When we truncate it to 32, 16, or 8 eigenvalues, the self-ranks degrade ($0.0\pm0.2$, $2.0\pm3.2$, and $8.6\pm9.1$, respectively).
This underscores the need for an appropriate estimate of the Hessian's eigenspace to maintain reliable task distinction. 
See~\aref{supp-sup: task_distinction} for details.
  
\subsubsection{Distinction of Normal and Noise Task Distributions}
\label{task_dist_distinction}
To validate Property 2, we prepare three training tasksets. Each set includes a training subpopulation that is less similar to the test tasks than the rest of the training tasks, achieved by replacing a subset of the base training taskset with tasks consisting of noise data.

\begin{figure*}[t]
	\begin{minipage}{0.46\linewidth}
		\centering
		\includegraphics[width=\linewidth, height=0.8\linewidth]
		{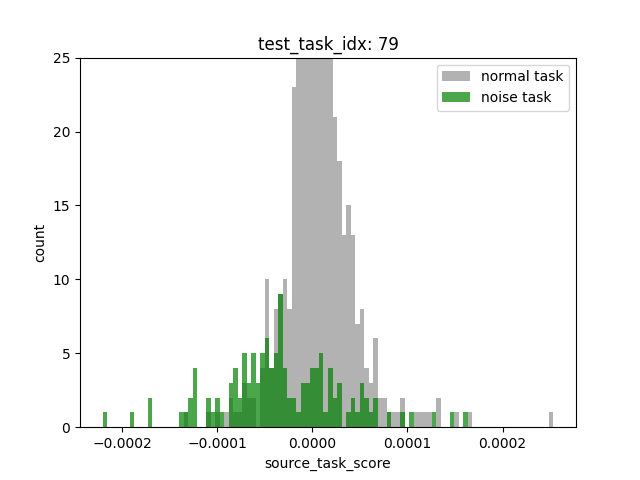}
		\caption{Example of influence score distribution for a single test task loss (synthetic dataset). 
		Green: noise training tasks. Gray: normal tasks 
		}
		\label{fig: proper_order}
	\end{minipage}%
	\hfill
	\begin{minipage}{0.54\linewidth}
		\centering
		\tblcaption{
			Experiment of distinguishing noise tasks and normal task distributions.
			MI and OM represent MiniImagenet and Omniglot respectively. The standard deviation $\sigma$ under the hypothesis of the random ordering is $\sqrt{0.5\times 0.5 \times 128} \sim 5.66$. 
			$\pm$ means the average and standard deviation across 5 runs of training.
		}
			\begin{tabular}{cccc}
				\toprule
				Meta-learner & Dataset & \multicolumn{2}{c}{Proper tests} \\
				{} &      {} & [count] &            [$\sigma$] \\
				\midrule
				MAML &      MI & 85.6$\pm$11.1 &  3.8$\pm$2.0 $\sigma$ \\
				MAML &      OM &  94.0$\pm$6.0 & 5.3$\pm$1.1 $\sigma$ \\
				Protonet &      MI &  95.6$\pm$5.3 &  5.6$\pm$0.9 $\sigma$ \\
				Protonet &      OM &  81.2$\pm$5.5 &  3.0$\pm$1.0 $\sigma$ \\
				\bottomrule
			\end{tabular}
			\label{tab: noise_distinction_small}
	\end{minipage}
\end{figure*}
\paragraph{Synthetic dataset.}
To examine the method with the exact form of \Eqref{eq:influence_ml_gen_param_mod}, we employ a lightweight network(3-layer, fully connected, 63 parameters) and create synthetic tasksets of clustered data points in a 2D plane that the network can learn easily.
Using Gaussian distributions, we first sample cluster centers around the origin of the plane, then the members of each cluster around its center.
Noise tasks consist of data points sampled around the origin and assigned random labels.
Figure~\ref{fig: proper_order} shows an example result of the network trained with 1024 training tasks contaminated by 128 noise tasks in the 3-ways-5-shots problem setting, and tested and explained with a single test task. 
We observe that the normal and noise tasks follow different distributions, and the former is scored higher, which aligns with our intuition. 
We refer to the positions of two distributions as ones in \textit {proper order} when they are in the intuitively natural order regarding their similarity to the test task, and a test that results in such a relation between the distributions as a \textit {test with the proper order} or a \textit{proper test}. 
We score the training tasks using the influence function over 128 test losses and observe that 113 tests give the proper order of the normal and noise task distributions, with their positions defined by the mean values of the subpopulations.
According to a two-sided binomial test with the null hypothesis of random ordering, this count exceeds the average count ($=64$) of the binomial distribution by $8.7 \sigma$, suggesting that the proposed method satisfies Property 2 statistically.

\paragraph{MiniImagenet and Omniglot.}
To validate Property 2 on more realistic tasks,  we train larger networks and adopt the approximation method in~\Eqref{eq: OPA} using MiniImagenet and Omniglot datasets (See \aref{supp-sup: runtime} for the effectiveness of the approximation method).
For each dataset, we use 8192 training tasks and replace the image tensors in 1024 of them with uniform noise tensors (noise images).
We train a network with 3-conv layers using MAML ($\sim$20k parameters) and a 4-conv-layer feature extractor using Protonet ($\sim$28k parameters). 
For each training setup, we evaluate the influence of the training tasks on 128 test loss values using \Eqref{eq:influence_ml_gen_perf} with the projected influence on the meta-parameters (\Eqref{eq:influence_ml_gen_param_mod}) and the GN matrix approximation (\Eqref{eq: OPA}).
Table~\ref{tab: noise_distinction_small} shows the number of proper tests in each training setting. Again, the binomial test suggests that our scoring method satisfies Property 2. See~\aref{supp-sup: noise_distinction} for details.

\subsubsection{Consistency with Semantic Similarity}
\label{task_dist_distinction_semantic}
\begin{figure}[t]
    \centering
    \includegraphics[width=\textwidth, height=0.25\textwidth]
        {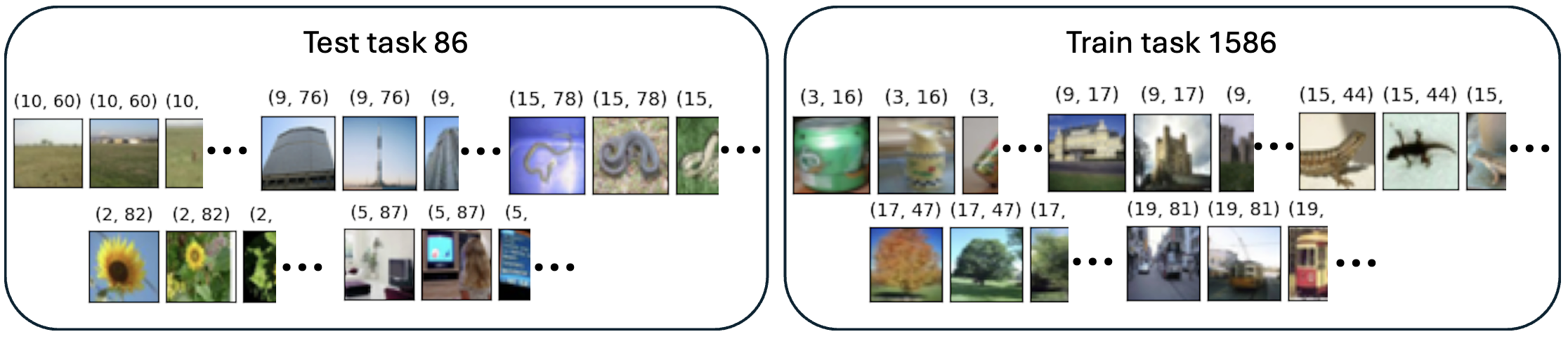}
    \caption{Examples of FC60 tasks. 
    The pair of labels on each image represents the semantic labels (superclass, subclass) obtained from CIFAR100. 
    }
    \label{fig: fc60_examples_small}

\end{figure}
\begin{table}[t]
	\centering
	\caption{
		Experiment of distinguishing training tasks with superclasses shared with test tasks from other training tasks.
		$\pm$ means the average and standard deviation across 1024 test tasks.
		The standard deviation $\sigma$ under the hypothesis of the random ordering is $\sqrt{0.5\times0.5\times1024}=16$.
		Expl. time means the time for generating explanation data, that is, the time spent for computing the 1024 test losses and the influence scores of the 8192 training tasks for each of them with a given $I^{\rm meta}$ for TLXML and the time spent for fine-tuning with the 1024 test tasks, computing the task embedding of each test task and the similarities with given 8192 training task embeddings for Task2Vec.
	}
	\label{tab: label_overlap_tlxml_t2v_small}
	\begin{tabular}{ccccccc}
		\toprule
		Method & Expl. time &    Overlap & \multicolumn{2}{l}{Train tasks} & \multicolumn{2}{l}{Test tasks}\\
		{}     & {}               & {}         &   total & filtered                      &   total & proper \\
		\midrule
		MAML+TLXML & 33sec &        1 & 8192 & 7522$\pm$314  &  1024 & 618 (+6.6$\sigma$) \\
		MAML+TLXML & 33sec &        2 & 8192 & 4781$\pm$718  & 1024 & 676 (+10.3$\sigma$) \\
		\midrule
		Protonet+TLXML & 56sec &   1 & 8192 & 7522$\pm$314  &  1024 & 658 (+9.1$\sigma$) \\
		Protonet+TLXML & 56sec &   2 & 8192 & 4781$\pm$718  & 1024 & 672 (+10.0$\sigma$) \\
		\midrule
		Task2Vec(8 steps) & 100sec &   1 & 8192 & 7522$\pm$314  &  1024 & 554 (+2.6$\sigma$) \\
		Task2Vec(8 steps)& 100sec &   2 & 8192 & 4781$\pm$718  & 1024 & 574 (+3.9$\sigma$) \\
		Task2Vec(64 steps) & 155sec &   1 & 8192 & 7522$\pm$314  &  1024 & 653 (+8.8$\sigma$) \\
		Task2Vec(64 steps)& 155sec &   2 & 8192 & 4781$\pm$718  & 1024 & 657 (+9.1$\sigma$) \\
		\bottomrule
	\end{tabular}
\end{table}

We also validate Property 2 using training and test tasks that share semantic information. 
We define a new image dataset, \textit{FC60}, in which each image is assigned hierarchical labels specifying the super- and sub-classes. The train/test splits 60 subclasses in total, and semantic similarities between train and test tasks are built in through 12 superclasses shared across the splits. 
This structure is realized by splitting the subclasses of each superclass in the training taskset of the FC100~\citep{oreshkin2018tadam}.
Figure~\ref{fig: fc60_examples_small} shows examples of a test and training task in FC60. 
We train a 3-conv-layer network using MAML and a 4-conv-layer feature extractor using Protonet with 8192 training tasks from FC60. 
Table~\ref{tab: label_overlap_tlxml_t2v_small} shows the number of proper tests with the training subpopulation defined by how many superclasses it shares with each test task, illustrating that TLXML distinguishes training subpopulations that share semantic properties with test tasks from the other training tasks.
The table also shows results obtained using Task2Vec~\citep{achille2019task2vec} as a baseline, with the same network structure as MAML. 
Task2Vec has the number of fine-tuning steps as a hyperparameter, and its larger value yields larger statistical significance but requires longer computation time. We have to increase it to around 64 to make Task2Vec on par with TLXML in terms of statistical significance; in that case, comparing the times for generating the explanation data,  we see that TLXML is 155 sec / 33 sec$\sim 5$, or 155 sec / 56 sec$\sim 3$ times faster than Task2Vec. 
The speed advantage of TLXML-based task similarity is a fruit of the efficient adaptation achieved through meta-learning.  
See~\ref{supp-sup: label_overlap} for details.

\subsection{Post-Convergence One-Step Update}
\label{one_step_update}
We examine the effect of the one-step update \Eqref{eq: one_step_update} 
from the convergence point of meta-learning.
We utilize 8192 training tasks from FC60 and train a network with 3-conv layers using MAML. 
Each training task is scored with the influence function for the average test loss across 1024 test tasks. 
\begin{table}[t]
\centering
\caption{Test accuracies after one-shot TLXML-guided update to a MAML-trained model with blocked tasks.
$\pm$ means the average and standard deviation across 5 runs of MAML training. 
Bold values outperform the MAML baseline (shown in the brackets) with a unpaired Welch two-sample \(t\)-test ($p<0.05$).
}
\label{tab: onestep_maml_small_xi}
\begin{tabular}{lccc}
\toprule
\textbf{test taskset} &
\textbf{\# tasks} &
Method & Accuracy\\
\midrule
\shortstack[l]{FC60}
& 2048 & $\xi=-8$ &0.659$\pm$0.009 \\
($0.663\pm0.004$)  &2048 & $\xi=-4$ & \textbf{0.674$\pm$0.007} \\
 &2048 & $\xi=-2$ & \textbf{0.672$\pm$0.007} \\
&2048 &  $\xi=-1$ & 0.669$\pm$0.004 \\
&2048 &  leave-out & 0.667$\pm$0.01 \\
\bottomrule
\end{tabular}
\end{table}
\begin{table}[t]
\centering
\caption{
Test accuracies after one-shot TLXML-guided update to MAML-trained models with blocked and enhanced tasks. See Table~\ref{tab: onestep_maml_small_xi} for notation.}
\label{tab: onestep_maml_small_dataset}
\begin{tabular}{lcc}
\toprule
\textbf{test taskset} &
\textbf{\# tasks} &
$\boldsymbol{\xi=-4}$ (\textbf{block}) \\
\midrule
\shortstack[l]{FC60~($0.663\pm0.004$)}
& 2048 & \textbf{0.674$\pm$0.007} \\
\shortstack[l]{FC100~($0.429\pm0.005$)}
 &2048 & \textbf{0.449$\pm$0.011} \\
\midrule
\textbf{test taskset} &
\textbf{\# tasks} &
$\boldsymbol{\xi=4}$ (\textbf{enhance}) \\
\midrule
\shortstack[l]{FC60~($0.663\pm0.004$)}
 &2048 & \textbf{0.676$\pm$0.008} \\
\shortstack[l]{FC100~($0.429\pm0.05$)}
&2048 & 0.429$\pm$0.010 \\
\bottomrule
\end{tabular}
\end{table}
Table~\ref{tab: onestep_maml_small_xi} shows the effects of blocking the 2048 lowest-score training tasks with one-step updates (and leave-out retraining for comparison) on the test accuracies. We observe that the test accuracies are improved by one-step update with suitable shift parameter values $\xi$, whereas leave-out retraining fails to yield a statistically significant gain.
Table~\ref{tab: onestep_maml_small_dataset} shows the effects of both blocking the 2048 lowest-score and enhancing the 2048 highest-score training tasks in two cases: test datasets with and without shared superclasses (FC60 test taskset and FC100 test taskset, respectively). We observe improvement in three of the four combinations of test settings. Although the case of enhancing the influence of highly-scored FC60 training tasks for the FC100 tests is an exception in the table, we can see that other values of $\xi$ boost the test performance in that setting. See~\aref{supp-sup: one_step_update} for details.

\section{Discussion and Conclusion}

This paper introduced TLXML, a method for quantifying the impact of meta-learning tasks.
We reduced its computational cost using the GN matrix approximation and handled the loss landscape's flat directions using the pseudo-inverse Hessian. Experiments suggest that TLXML generates adequate explanation data, helping actionable insights and improved performance in downstream tasks.

Extending TLXML beyond classification tasks, e.g., regression and reinforcement learning (RL), is expected to be straightforward, as its formulation in terms of gradients is almost agnostic to differences across learning objectives. 
In addition, because TLXML provides a way to define task similarity, future work should explore its relation to general notions, such as out-of-distribution awareness and domain adaptation.

One limitation is the assumption of a local minimum of the loss function in the definitions of the influence functions.
Future work could aim to design influence functions compatible with early stopping techniques.
Another is that the experiments were done only on MAML and Prototypical Networks.
Although they are representative methods for expressing meta-knowledge using initial model parameters and an embedding function, respectively, it is worth strengthening the evidence through experiments with other types of meta-learning.
For example, TLXML can be applied straightforwardly to feed-forward weight-generation methods exemplified by Learnet \citet{bertinetto2016learning} and HyperTransformer \citet{zhmoginov2022hypertransformer}.
We also note that the current evaluation is purely quantitative, using standard benchmark datasets; investigating how non-experts would leverage TLXML is a worthwhile direction for future work.

\section*{Acknowledgments}

The authors are grateful to Sheila Schoepp for helpful discussions at the early stages of this work and warm assistance with improving and formatting draft versions.

\bibliographystyle{splncs04}
\bibliography{bib/all,bib/task_relatedness.bib,bib/meta_learning.bib, bib/ecmlpkdd.bib}

\clearpage
\appendix
\PushLabel{supp-}
\section{Related work details}
\label{sup: related_work}

\paragraph{Influence functions for supervised learning.}
Research on Influence functions for modern supervised machine learning initiated by ~\citet{koh2017understanding} includes various application areas, such as crafting/detecting adversarial training examples~\citep{cohen2020detecting}, quantifying model uncertainty~\citep{alaa2020discriminative}, improving toxicity classifiers ~\citep{han2020fortifying}, interpreting feature spaces~\citep{chhabra2024data}, and analyzing to NLP models~\citep{han2020explaining}, including large language models~\citep{grosse2023studying}.
This work extends the influence functions proposed for supervised learning~\citep{koh2017understanding} to meta-learning settings.
As in supervised learning~\citep{khanna2019interpreting,Barshan2020relatif}, we will see Fisher information metric offers a useful geometric perspective on influence functions for meta-learning. We will also see that it plays a role peculiar to meta-learning in handling a computational cost challenge.
We also note that the non-convexity of loss functions is argued to lead to erroneous influence functions in supervised learning~\citep{basu2021influence}. We will see that the same holds in meta-learning and address it by generalizing the definition of the influence functions.

\paragraph{Explainable AI (XAI) for meta-learning.} 
Naturally, XAI methods that are agnostic to the learning process are applicable for explaining inference in meta-learning. 
Although this area is still in its early stages,
several research examples exist~\citep{woznica2021towards,sijben2024function,shao2023effect}.
With a motivation close to ours, \citet{woznica2021towards} quantified the importance of high-level characteristics of a dataset, such as the size, the number of features, and the number of classes.
However, we emphasize that the strength of meta-learning lies in its ability to generalize across \textit{tasks}, and understanding the impact of training tasks is crucial for evaluating a model's adaptability.

\paragraph{Impact of training data.} 
The impact of training data is frequently discussed in relation to the robustness of machine learning models~\citep{khanna2019interpreting,ribeiro2016should}. 
This topic has also been explored in the context of meta-learning, such as creating training-time adversarial attacks via meta-learning~\citep{zugner2019adversarial,xu2021yet}, training robust meta-learning models by exposing models to adversarial attacks during the query step of meta-learning~\citep{goldblum2020adversarially}, and data augmentation for enhancing the performance of meta-learning algorithms~\citep{ni2021data}.
However, influence functions have not been utilized for those purposes to date.

\paragraph{Task-relatedness.}
Apart from the breadth of potential applications of TLXML, one restricted application is quantifying the relatedness between training and test tasks.
Task-relatedness is a frequently discussed subject in multi-task learning and meta-learning.
Known quantification methods include ones that use distance or divergence measures between data distributions in tasks~\citep{shui2019principled}, the loss gradient for each task~\citep{chen2018gradnorm,yu2020gradient}, transferability from one task to another~\citep{zamir2018taskonomy,phuoc2022task}, task-embeddings obtained by fine-tuning a model with each task~\citep{achille2019task2vec}, and annotations of additional task attributions~\citep{hu2023understanding}.
Compatibility with the motivation of this work requires that the quantification be achieved without access to the training tasks or additional annotation in user environments.
We will compare TLXML-based task similarity with Task2Vec~\citep{achille2019task2vec}, as a well-established method that satisfies these requirements.

\paragraph{Machine Unlearning (MU).}
Removing the influence of training data from a trained model represents a valuable application of influence functions beyond increasing the interpretability of machine learning models. 
Aiming to handle privacy concerns in supervised learning, ~\citet{cao2015towards} introduced MU as a method of forgetting a training point 
without incurring the cost of retraining the model from scratch.
Although MU was first proposed outside the scope of machine learning interpretability, subsequent research established close connections to influence functions.
\citet{guo2020certified} discussed certified removal of data influence, noting that the influence function approximates the Newton method update corresponding to deleting a data point.
\citet{Golatkar2020eternal,Golatkar2021mixed} re-derived influence functions from an information-theoretic argument.
See \citet{li2025machine} for a recent review that includes those progresses.
In this paper, we show that TLXML extends MU to meta-learning and can be used to both block and amplify the influence of training tasks.

\paragraph{Extension of TLXML.}
After the core ideas of this work were presented in earlier preprints, a related work \citet{ren2025evaluating} appeared.
After revisiting the argument in this work on the definition of task-level influence function and its computational barriers, it suggests replacing the Hessian with a diagonal matrix constructed from the loss-gradient magnitude vector. Although its theoretical justification is awaited, the proposed proxy for the influence function has been empirically shown to be useful for identifying harmful tasks for MAML models.
Another notable aspect not covered by this work is the extension to the influence of individual data points on each task, motivated by the goal of improving meta-training. In contrast, the motivation for this work is to provide users with concise explanations aligned with user-level abstraction, prioritizing task-level explanations over instance-level explanations.

\clearpage
\section{Method details}
\subsection{Meta-learning Algorithms}
\label{sup: meta_learners}
We present the explicit forms of the adaptation algorithm $\mathcal A$ used for MAML and Protonet.

In MAML, the initial values $\theta_0$ of the network weights 
serve as the meta-parameters, 
and  the adaptation $\mathcal{A}$ represents a one-step gradient descent update of the weights with a fixed learning rate $\alpha$:
\begin{align*}
\mathcal{A}(\mathcal{D}^{(i)\rm{train}}, \mathcal{L}^i, \theta_0) =
\theta_{0}-\alpha\left.\partial_{\theta}\mathcal{L}^{i}(\mathcal{D}^{(i)\rm{train}},f_{\theta})\right|_{\theta=\theta_{0}}
.
\end{align*}

In Protonet, the meta-parameters are the weights of a feature extractor $f_\theta$
and the adaptation $\mathcal A$ computes the feature centroid of the data points belonging to each class in the support, without involving a loss function.
Formally, it can be expressed as a function vector that passes the weights $\theta$ without any modification and calculates the feature centroid $c_k$ for 
each class $k$ based on the support set $S_k \in \mathcal D$:
\begin{align*}
\theta 
&= 
\mathcal{A}_\theta(\mathcal{D}^{(i)\rm{train}}, \mathcal{L}^i, \theta),\\
c_k
&=
\mathcal{A}_k (\mathcal{D}^{(i)\rm{train}}, \mathcal{L}^i, \theta)
 = \frac{1}{|S^{(i)}_k|} \sum_{(x,y)\in S_k^{(i)}} f_\theta (x).
\end{align*}
The class prediction for each data point $x$ 
is given by the following weighted distances to the centroids:
\begin{align*}
P_\theta (l|x) = \frac{\exp d(f_\theta (x), c_l)}{\sum_k \exp d(f_\theta (x), c_k)},
\end{align*}
where $d$ is a distance measure(e.g. Euclidean distance).

\subsection{Implicit differentiation}
\label{sup: implicit_differentiation}
The second equation in each of
\mEqref{eq:inf_func_def},
\mEqref{eq:influence_ml_gen_params},
and \mEqref{eq:influence_ml_gen_param_mod}
are derived immediately from the following property.

\begin{prop}
If a vector $\hat\theta_\epsilon$ is parametrized by a scalar parameter $\epsilon$ in a way that the local maximum or the local minimum condition of a second-order differentiable function $L(\theta,\epsilon)$ is satisfied for each value of $\epsilon$,  
then
the derivative of $\hat\theta_\epsilon$ with respect to $\epsilon$ satisfies
\begin{align}
\left.
\frac{\partial^{2}L(\theta,\epsilon)}{\partial\theta\partial\theta}
\right|_{\theta=\hat\theta_\epsilon}
\frac{d\hat{\theta}_{\epsilon}}{d\epsilon} & =
\left.
- \frac{\partial L\left(\theta,\epsilon\right)}{\partial\theta\partial\epsilon}
\right|_{\theta=\hat{\theta}_{\epsilon}}.\label{eq:implicit_diff_formula}
\end{align}
\end{prop}
\begin{proof}
The proof is done almost trivially by differentiating the local maximum or minimum condition 
\begin{align}
\left.\frac{\partial L\left(\theta,\epsilon\right)}{\partial\theta}\right|_{\theta=\hat{\theta}_{\epsilon}} &=0
\label{eq: min_max_cond}
\end{align}
with respect to $\epsilon$ and apply the chain rule.
\qedsymbol
\end{proof}
Note that if the matrix $\partial\partial L$ in \Eqref{eq:implicit_diff_formula} is invertible, we can solve the equation to obtain the $\epsilon$-derivative of $\hat\theta_\epsilon$.
Note also that we do not assume $\hat\theta_\epsilon$ to be the unique solution of \Eqref{eq: min_max_cond} and \Eqref{eq:implicit_diff_formula} is true for any parametrization of $\theta$ with $\epsilon$ that satisfies \Eqref{eq: min_max_cond}.

\subsection{Task Grouping}
\label{sup: task_grouping}
In some cases, task-level abstraction is insufficient, and explanations based on task groups are more suitable. 
This requirement arises when some tasks in the training taskset are too similar to each other for human intuition.
For example, when an image recognition model is trained with task augmentation(see~\citet{ni2021data} for terminology of data augmentation for meta-learning), e.g., flipping, rotating, or distorting the images in original tasks, the influence of each deformed task is not of interest; rather, the influence of the task group generated from each original task is of interest.

We extend the definition of influence functions 
to the task-group level 
by considering a common perturbation $\epsilon$ in the losses of tasks within a task group
$\mathcal{G}^J{=}\{\mathcal{T}^{j_0},\mathcal{T}^{j_1},\cdots\}$:
\begin{align}
\hat{\omega}_{\epsilon}^{J}
&=
\mathop{\underset{\omega}{\arg\min}}\frac{1}{M}
\left[ 
\sum_{i=1}^{M}\mathcal{L}^{i}(\mathcal{D}^{(i){\rm test}},f_{\hat{\theta}^{i}})
+\epsilon\sum_{\mathcal{T}^j\in \mathcal{G}^J}
\mathcal{L}^{j}(\mathcal{D}^{(j){\rm test}},f_{\hat{\theta}^{j}})
\right] \\
 & \text{with}\ 
 \hat{\theta}^{i}{=}\mathcal{A}(\mathcal{D}^{(i){\rm train}},\mathcal{L}^i,\omega), \nonumber
\end{align}
which modifies the influence function in
\mEqref{eq:influence_ml_gen_params} as
\begin{align}
I^{{\rm meta}}(J)\overset{{\rm def}}{=} & 
  \left.\frac{d\hat{\omega}_{\epsilon}^{J}}{d\epsilon}\right|_{\epsilon=0}
= \sum_{\mathcal{T}^j\in \mathcal{G}^J} I^{{\rm meta}}(j).
\label{eq1:influence_ml_gen_gr_params}
\end{align}
\mEqref{eq:influence_ml_gen_adpt} and \mEqref{eq:influence_ml_gen_perf}
are only affected by replacing the task index $j$ with the task-group index $J$. The derivation of \Eqref{eq1:influence_ml_gen_gr_params} is done by directly applying the argument in \ref{sup: implicit_differentiation}. 
The linear structure of influence functions inherently arises from the linear structure of the loss function; therefore, it also holds in supervised learning, and, based on that property, influences of batched data points are studied in ~\citet{koh2019accuracy}.

\subsection{Third-order tensors in influence functions} \label{sup: explicit_forms}
Here, we explain how the computational cost of $\mathcal{O}(pq^2)$ arises in evaluating the influence function in \mEqref{eq:influence_ml_gen_params}. This is due to the third-order tensors appearing in the intermediate process of evaluating the Hessian:  
\begin{align}
H = & \frac{1}{M}\sum_{i=1}^{M}
\frac{\partial^{2}\mathcal{L}^{i}(\mathcal{D}^{(i)\rm{test}},f_{\hat{\theta}^{i}(\omega)})}{\partial\omega\partial\omega}\\
= &
\frac{1}{M}\sum_{i=1}^{M}\left[ 
\left(\frac{\partial\hat{\theta}^{i}\left(\omega\right)}{\partial\omega}\right)^{{\rm T}}
\left.
\frac{\partial^2\mathcal{L}^{i}
(\mathcal{D}^{(i)\rm{test}},f_{\theta})
}{\partial\theta\partial\theta}
\right|_{\theta=\hat\theta^{i}(\omega)}
\frac{\partial\hat{\theta}^{i}\left(\omega\right)}{\partial\omega}
\right.\nonumber\\
& +
\left.\left.
\frac{\partial\mathcal{L}^{i}
(\mathcal{D}^{(i)\rm{test}},f_{\theta})
}
{\partial\theta}
\right|_{\theta=\hat\theta^{i}(\omega)}
\frac{\partial^{2}\hat{\theta}^{i}\left(\omega\right)}{\partial\omega\partial\omega}
\right]
\end{align}
where $\hat{\theta}^{i}=\mathcal{A}(\mathcal{L}^{(i){\rm train}},\mathcal{L}^i,\omega)$. 
Because $\hat\theta^i$ and $\omega$ have dimensions $p$ and $q$ respectively, the second-order derivative
$\partial\partial \hat\theta^i$
in the last term is the third-order tensor of $pq^2$ elements. This tensor also appears in the evaluation of the second-order derivative of the network output with respect to $\omega$. In the case of MAML, this tensor is in the form of a third-order derivative:
\begin{align*}
\frac{\partial^{2}\hat{\theta}^{i}\left(\theta_0\right)}{\partial\theta_0\partial\theta_0}
&=
\frac{\partial^{2}}{\partial\theta_0\partial\theta_0}
\mathcal{A}(\mathcal{D}^{(i){\rm train}},\mathcal{L}^i,\theta_0)\\
&=
-\alpha \frac{\partial^3}{\partial\theta_0\partial\theta_0\partial\theta_0}
\mathcal{L}^i\left(\mathcal{D}^{(i){\rm train}}, f_{\theta_0}\right).
\end{align*}

\subsection{Relations Among KL-divergence, Cross-Entropy, and Fisher Information Metric}\label{sup: OPA_proof}
For the reader's convenience, we present some basic facts about the approximation method argued in Section~\mref{hessian_approximation}. 

\paragraph{Variant expressions of Hessian.}
The cross-entropy $L$ between two probability distributions
$P\left(X|\omega^{*}\right)$, $P\left(X|\omega\right)$
parametrized by $\omega^*$ and $\omega$,
is equivalent to the Kullback–Leibler (KL) divergence up to a $\omega$-independent term:
\begin{align*}
D_{KL}\left(P\left(X|\omega^{*}\right),P\left(X|\omega\right)\right)
=&
L\left(P\left(X|\omega^{*}\right),P\left(X|\omega\right)\right) \\
&
+\int P(X|\omega^{*})\log P(X|\omega^{*})dX.
\end{align*}
Therefore, the second-order derivatives of them with respect to $\omega$ are identical:
\begin{align*}
\frac{\partial^2}{\partial \omega \partial \omega} D_{KL}
&=
\frac{\partial^2}{\partial \omega \partial \omega} L.
\end{align*}
Furthermore, considering the Taylor expansion of $D_{KL}$ with
$\Delta \omega \equiv \omega{-}\omega^*$:
\begin{align*}
& 
D_{KL}\left(P\left(X|\omega^*\right),P\left(X|\omega\right)\right) \\
& \sim
- \sum_\mu \Delta\omega^\mu 
\left[ \int \partial_\mu P\left(X|\omega^*\right) dX \right]\\
& +
\frac{1}{2}\sum_{\mu\nu}\Delta\omega^\mu\Delta\omega^\nu
\left[ 
\int -\partial_{\mu}\partial_{\nu} P\left(X|\omega^*\right)
+ \frac{\partial_{\mu} P\left(X|\omega^*\right)\partial_{\nu}P\left(X|\omega^*\right)^2}{P\left(X|\omega^*\right)} dX
\right]\\
& =
\frac{1}{2} \sum_{\mu\nu} \Delta\omega^\mu\Delta\omega^\nu
\mathrm{E}_{X\sim P(X|\omega^{*})}\left[\partial_{\mu}\log P\left(X|\omega^{*}\right)\partial_{\nu}\log P\left(X|\omega^{*}\right)\right]\\
& \equiv
\frac{1}{2}\sum_{\mu\nu} g_{\mu\nu}\left(\omega^{*}\right)\Delta\omega^{\mu}\Delta\omega^{\nu},
\end{align*}
where $g_{\mu\nu}$ is the Fisher information metric, we see that
\begin{align}
\left.
\frac{\partial^2}{\partial \omega_\mu \partial \omega_\nu} D_{KL}
\right|_{\omega=\omega^*}
&=
\left.
\frac{\partial^2}{\partial \omega_\mu \partial \omega_\nu} L
\right|_{\omega=\omega^*}
=
g_{\mu\nu}\left(\omega^*\right) \label{eq: ddK_ddL_g}.
\end{align}

\paragraph{Approximations by empirical sums.}
Let us consider the case that $X=(x, c)$ is the pair of a network input $x$ and a class label $c$ and the conditional distribution $P(c|x)$ is the composition of softmax function $\sigma$ and the network output $\boldsymbol{y}\equiv f_\omega(x)$. Then, the Fisher information metric at $\omega^*$ can be written as
\begin{align*}
g_{\mu\nu}\left(\omega^{*}\right) 
&=
\mathrm{E}_{X\sim P(X|\omega^{*})}
\left[
\partial_{\mu}\log P\left(X|\omega^{*}\right)\partial_{\nu}\log P\left(X|\omega^{*}\right)
\right]\\
& =
 \mathrm{E}_{(c,x)\sim P_{\omega^*}\left(c|x\right)P(x)}
 \left[
\left. 
\partial_{\mu}\log\left(P_{\omega}\left(c|x\right)P(x)\right)
\partial_{\nu}\log\left(P_{\omega}\left(c|x\right)P(x)\right)
 \right]
 \right|_{\omega=\omega^{*}}\notag\\
& =
 \mathrm{E}_{(c,x)\sim P_{\omega^*}\left(c|x\right)P(x)}
 \left[
\left. 
 \partial_{\mu}\log\left(P_{\omega}\left(c|x\right)\right)
 \partial_{\nu}\log\left(P_{\omega}\left(c|x\right)\right)
 \right]
 \right|_{\omega=\omega^{*}}.\notag
\end{align*}
Assuming that sampled data accurately approximate the distributions, we obtain the expressions of the Fisher information metric in the form of an empirical sum:
\begin{align}
g_{\mu\nu}\left(\omega^{*}\right)
 & \sim\left.\sum_{ni}\sigma_{i}\left(\boldsymbol{y}_{n}\right)\left[\partial_{\mu}\log\left(\sigma_{i}\left(\boldsymbol{y}_{n}\right)\right)\partial_{\nu}\log\left(\sigma_{i}\left(\boldsymbol{y}_{n}\right)\right)\right]\right|_{\omega=\omega^{*}}\notag\\
 & =\left.
 \sum_{nijk}\sigma_{i}\left(\boldsymbol{y}_{n}\right)
 \left(\delta_{ik}-\sigma_{k}\left(\boldsymbol{y}_{n}\right)\right)
 \left(\delta_{ij}-\sigma_{j}\left(\boldsymbol{y}_{n}\right)\right)
 \frac{\partial y_{nk}}{\partial\omega_{\mu}}
 \frac{\partial y_{nj}}{\partial\omega_{\nu}}
 \right|_{\omega=\omega^{*}}\notag\\
 &=\left.
 \sum_{nkj}\sigma_{k}
 \left(\boldsymbol{y}_{n}\right)
 \left(\delta_{kj}-\sigma_{j}\left(\boldsymbol{y}_{n}\right)\right)
 \frac{\partial y_{nk}}{\partial\omega_{\mu}}\frac{\partial y_{nj}}{\partial\omega_{\nu}}
 \right|_{\omega=\omega^{*}}\label{eq: g_emp_sum}
\end{align}
where $\sigma_i(\boldsymbol{y}_n)$ is the probability of class $i$ predicted by the network ourtput $\boldsymbol{y}_n$ and softmax function $\sigma_i$ for the sample $x_n$.
To express the Hessian in a similar way, we denote the target vector of $x_n$ as $t_{ni}$. Then, the Hessian of the cross-entropy at $\omega$ (not $\omega^{*}$) can be written as
\begin{align}
& \frac{\partial^2}{\partial \omega \partial \omega} L \notag\\
&=
-\frac{\partial^2}{\partial \omega \partial \omega}
\sum_c
\int P_{\omega^*}\left(c|x\right)P(x)
\log \left[ P_{\omega}\left(c|x\right)P(x)\right]
dx \notag\\
& \sim
-\frac{\partial^2}{\partial \omega \partial \omega}
\sum_{ni}t_{ni}
\log \left[ P_{\omega}\left(i|x_n\right)P(x_n)\right] \notag\\
& = 
-\frac{\partial^2}{\partial \omega \partial \omega}
\sum_{ni}t_{ni}
\log \sigma_i(\boldsymbol{y_n}) \notag\\
&=
-\frac{\partial}{\partial \omega}
\sum_{nik}t_{ni}
\left[
\delta_{ik} - \sigma_k(\boldsymbol{y_n})
\right]
\frac{\partial y_{nk}}{\partial \omega}\notag\\
& = 
\sum_{nijk}t_{ni}
\sigma_k(\boldsymbol{y_n})\left(\delta_{kj} - \sigma_j(\boldsymbol{y_n})\right)
\frac{\partial y_{nk}}{\partial \omega}\frac{\partial y_{nj}}{\partial \omega} 
-
\sum_{nik}t_{ni}
\left[
\delta_{ik} - \sigma_i(\boldsymbol{y_n})
\right]
\frac{\partial^2 y_{nk}}{\partial \omega \partial \omega},\notag
\end{align}
and because $\boldsymbol{t_n}$ is a one-hot vector ($\sum_i t_{ni} = 1$), this equals to
\begin{align}
&=
\sum_{njk}
\sigma_k(\boldsymbol{y_n})\left(\delta_{kj} - \sigma_j(\boldsymbol{y_n})\right)
\frac{\partial y_{nk}}{\partial \omega}\frac{\partial y_{nj}}{\partial \omega} 
-
\sum_{nik}t_{ni}
\left[
\delta_{ik} - \sigma_i(\boldsymbol{y_n})
\right]
\frac{\partial^2 y_{nk}}{\partial \omega \partial \omega}\notag\\
&=
g(\omega)
-
\sum_{nik}t_{ni}
\left[
\delta_{ik} - \sigma_i(\boldsymbol{y_n})
\right]
\frac{\partial^2 y_{nk}}{\partial \omega \partial \omega}. \label{eq: ddL}
\end{align}
In the last equation, we have used the expression for the Fisher information metric \Eqref{eq: g_emp_sum}. The second-to-last equation proves \mEqref{eq: preOPA} in the main text. 
From the entire derivation of \Eqref{eq: ddL}, we see that when the distributions $P(X|\omega^*)$ and $P(X|\omega)$ 
are accurately approximated by the training data samples 
and the model's outputs, respectively, 
the first term in \mEqref{eq: preOPA} in the main text
becomes equivalent to the Fisher information metric evaluated at $\omega$.
In addition to that condition, if $\omega$ is near to $\omega^*$, comparing \Eqref{eq: ddK_ddL_g} and \Eqref{eq: ddL} leads to that the second term of \Eqref{eq: ddL} drops, and the Hessian is well approximated by the Fisher information metric evaluated at $\omega$, that is, $g_{\mu\nu}(\omega)$.

We should note that the above argument is agnostic about whether $\omega$ are network weights trained in a supervised learning setting or meta-parameters trained in a meta-learning setting. 
The difference lies only in the internal structure of the quantities involved. 
For instance, in meta-learning, data distributions internally have integration over the tasks $\mathcal{T}$ and their dependency on the meta-parameters $\omega$ is given through the adaptation of the weights $\hat\theta$ :
\begin{align*}
P(x)
&= 
\int P\left( x | \mathcal{T}\right)P\left(\mathcal{T}\right) d\mathcal{T} \\
P_{\omega}\left(c|x\right)
&=
\int P\left(c | x, \omega, \mathcal{T}\right) P\left(\mathcal{T}\right) d\mathcal{T}
=
\int 
P\left(c | x, \hat\theta\left(\omega, \mathcal{T}\right)\right)
P\left(\mathcal{T}\right) d\mathcal{T}
\end{align*}

\subsection{Implementation}\label{sup: implementation}
\begin{figure}[t]
\begin{algorithm}[H]
	\caption{Orthogonalized columns in GN matrix}
	\label{alg: calc_opa}
    \begin{algorithmic}[1]
    \Require
    Model $f$,
    Adaptation algorithm $\mathcal{A}$
    \Statex \hspace{\algorithmicindent}
    Training taskset
    $D=\{(\mathcal{D}^{(i)\rm{sup}}, \mathcal{D}^{(i)\rm{val}})\}_{i=1}^M$
    \Statex \hspace{\algorithmicindent}
    Trained meta-parameters $\hat{\omega}$
    \Statex \hspace{\algorithmicindent}
    \# output columns $N_{\mathrm{orth}}$
    \Statex \hspace{\algorithmicindent}
    Vector buffer size $N_{\mathrm{max}}$
    \Ensure Matrix with orthogonal columns $V_{\rm{orth}}$
    \State $V_{\rm{orth}}\gets \text{empty matrix}$ 
    \ForAll {$(\mathcal{D^{\rm{sup}}}, \mathcal{D^{\rm{val}}})\gets D$}
    \State
    $\hat{\theta} \gets \mathcal{A}(\mathcal{D}^{\rm{sup}}, \hat{\omega})$
    \ForAll {$(x, t) \gets \mathcal{D^{\rm{val}}}$}
    \State
    $\bm{y} \gets f(x;\hat{\theta})$,
    $\bm{\sigma} \gets \mathrm{Softmax} (\bm{y})$
    \State
    $
    \bm{v} \gets
     \partial \bm{y} / \partial\omega
     |_{\omega=\hat\omega}
    $
    \State
    $
    \bm{v} \gets
    \mathrm{Sqrt}(
      \mathrm{DiagonalMatrix}(\mathbf{\bm{\sigma}})
      - \mathbf{\bm{\sigma}} \mathbf{\bm{\sigma}}^{\mathrm{T}}
      ))
      \cdot
     \bm{v}
    $
    \State $V_{\rm{orth}} \gets
    \mathrm{HorizontalStack}(V_{\rm{orth}}, \bm{v})$
    \EndFor
    \If{\# columns of $V_{\rm{orth}} > N_{\mathrm{max}}$}
    \State  $V_{\rm{orth}} \gets
    \mathrm{OrthogonalColumns}(V_{\rm{orth}}, N_{\mathrm{orth}})$
    \EndIf
    \EndFor
    \State  $V_{\rm{orth}} \gets
    \mathrm{OrthogonalColumns}(V_{\rm{orth}}, N_{\mathrm{orth}})$
    \State\Return $V_{\rm{orth}}$
    \end{algorithmic}
\end{algorithm}
\end{figure}
\begin{figure}[t]
\begin{algorithm}[H]
	\caption{Orghogonal Columns}
	\label{alg: select_col}
    \begin{algorithmic}[1]
    \Require Matrix $V$, \# output columns $N_{\mathrm{orth}}$
    \Ensure
    \Statex Matrix with orthogonal columns $V_{\mathrm{orth}}$
    \State Compute orthogonal matrix $O$, diagonal matrix $\Lambda$,
    such that $V^{\mathrm{T}}V = O\Lambda O^{\mathrm{T}}$  
    \State List of eigenvalues $\bm{\lambda} \gets \mathrm{DiagonalPart}(\Lambda)$
    \State Extract a set of indices $L = \{i_1,\cdots,i_{N_{\mathrm{orth}}}\}$ such that $\bm{\lambda}[i_1],\cdots,\bm{\lambda}[i_{N_{\mathrm{orth}}}]$ is the largest $N_{\mathrm{orth}}$ eigenvalues
    \State $V_{\rm{orth}} \gets VO[:, L]$
    \State \Return $V_{\rm{orth}}$
    \end{algorithmic}
\end{algorithm}
\begin{algorithm}[H]
	\caption{Orthogonal Columns in pseudo-inverse GN matrix}
	\label{alg: calc_pseudo_opa}
    \begin{algorithmic}[1]
    \Require Matrix with orthogonal columns
    $V_{\mathrm{orth}}$ of size $n\times m$
    \Ensure
    \Statex Matrix with orthogonal columns $V_{\mathrm{orth}}^{+}$
    \State $V_{\rm{orth}}^{+} \gets$ empty matrix
    \ForAll{$j \gets [1,\cdots, m]$}
    \State $\bm{v}\gets V_{\rm{orth}}[:, j]$
    \If{$|\bm{v}| > 0$}
    \State $\bm{v}\gets \bm{v}/|\bm{v}|^2$
    \EndIf
    \State
    $V_{\rm{orth}}^{+} \gets \mathrm{HorizontalStack}(V_{\rm{orth}}^{+}, \bm{v})$
    \EndFor
    \State \Return $V_{\rm{orth}}^{+}$
    \end{algorithmic}
\end{algorithm}
\end{figure}

We comment on the implementation for computing the approximated Hessian \mEqref{eq: OPA}  and its pseudo-inverse $H^+$ in \mEqref{eq:influence_ml_gen_param_mod} in the main text.
Here we restate the GN-matrix approximation of the Hessian:
\begin{align}
H
&\sim
\sum_{nj}\left(\mathbf{V}\right)_{\mu(nj)}\left(\mathbf{V}^{{\rm T}}\right)_{(nj)\nu} 
=
VV^T
\label{seq: OPA}
\end{align}
where we define the matrix $V$ as the horizontal concatenation of $q$ dimensional column vectors $\mathbf{V}_{(nj)}$. The block specified by $n$, associated with each query in each training task, satisfies
\begin{align*}
&
\sum_{j}
\left(\mathbf{V}\right)_{\mu(nj)}\left(\mathbf{V}^{{\rm T}}\right)_{(nj)\nu}
 =
 \sum_{jk}\sigma_{k}\left(\boldsymbol{y}_{n}\right)\left(\delta_{kj}-\sigma_{j}\left(\boldsymbol{y}_{n}\right)\right)\frac{\partial y_{nk}}{\partial\omega}\frac{\partial y_{nj}}{\partial\omega}
,
\end{align*}
where $\mu, \nu$ are the meta-parameter indices, $n$ is the training task and query index, and $k,j$ are the output tensor indices.  
Although the Gauss-Newton matrix approximation avoids the $\mathcal{O}(pq^2)$ computational cost for handling a network with $p$ model parameters and $q$ meta-parameters, the large size of $H$, i.e. $q\times q$, leads to a high storage/memory cost. 
We can mitigate it by keeping only $V$ and conducting all manipulation without evaluating the matrix product in \Eqref{seq: OPA}. However, when the training taskset is large, the number of columns of $V$ itself is also large. 
\footnote{For example, if the network has $q=$100k meta-parameters, the training taskset has $M=$1000 tasks, and each task is a 5-way-5-shot problem($C=5$ output size, $Q=5\times5$ queries per task), then the number of matrix elements is 
$qCQM{=}10^5{\times}5{\times}25{\times}1000{\sim}10^{10}$, 
which makes it challenging to use floating-point numbers with 16, 32, or 64-bit precision.}
To resolve this issue, we orthogonalize the columns to remove redundancy in their linear dependencies.
As explained below, this can be done by sequentially orthogonalizing column vectors (Algorithm~\ref{alg: calc_opa}), thereby avoiding the costs of accumulating all terms and orthogonalizing a large set of vectors at once.
We also note that, in conducting the experiments in this paper, methods for further approximating the GN-matrix, such as K-FAC~\citep{martens2015optimizing}, are not necessary.

First, we show that removing columns from a matrix by the orthogonalization of Algorithm~\ref{alg: select_col} preserves the matrix product.
\begin{prop}\label{prop: OPA_reduction}
Let $V$ and $N_{\mathrm{orth}}$ be the input to Algorithm \ref{alg: select_col}, and $V_{\mathrm{orth}}$ be its output. If $N_{\mathrm{orth}}$ is greater than or equal to the number of linearly independent columns of $V$, then
\begin{align*}
VV^{\mathrm{T}}
=
V_{\mathrm{orth}} V_{\mathrm{orth}}^{\mathrm{T}}.
\end{align*}
\end{prop}

\begin{proof}
We use the notations in Algorithm\ref{alg: select_col}. Let $N$ be the number of columns of $V$. From $O^{\mathrm{T}}V^{\mathrm{T}}VO=diag(\lambda_1,\cdots,\lambda_N)$, we see the columns of $VO$ are orthogonal to each other:
\begin{align*}
\bm{v}_i\cdot \bm{v}_j = 0 \text{ for } i\ne j,
VO \equiv [\bm{v}_1, \cdots, \bm{v}_N]
\end{align*}
and their squared norms are the eigenvalues: $\bm{v}_i\cdot \bm{v}_i = \lambda_i$.
Noting $VV^{\mathrm{T}}$ is positive semi-definite and $N_{\mathrm{orth}} \ge \mathrm{rank} VV^{\mathrm{T}}$, we see that $\lambda_i = 0$ and hence $\bm{v}_i = \bm{0}$ for $i\notin L$. Thus we obtain
\begin{align*}
VV^T
&= VOO^TV^T
= \sum_{i=0}^N \bm{v}_i \bm{v}_i^T
= \sum_{i\in L} \bm{v}_i \bm{v}_i^T
= V_{\mathrm{orth}} V_{\mathrm{orth}}^{\mathrm{T}}. 
\end{align*}
\qedsymbol
\end{proof}

Let us consider a split of columns of a matrix $V = [V^{\prime}, V^{\prime\prime}]$. A proposition similar to the above also holds in the case that Algorithm~\ref{alg: select_col} is applied to the submatrix $V^{\prime}$. 
\begin{prop}\label{prop: OPA_partial_reduction}
Let $V^{\prime}$ and $N_{\mathrm{orth}}$ be the input to Algorithm \ref{alg: select_col}, and $V'_{\mathrm{orth}}$ be its output. If $N_{\mathrm{orth}}$ is greater than or equal to the number of linearly independent columns of $V$, then
\begin{align*}
VV^{\mathrm{T}}
=
V_{\mathrm{orth}}^{\prime} V_{\mathrm{orth}}^{\prime\mathrm{T}}
+
V^{\prime\prime} V^{\prime\prime\mathrm{T}}.
\end{align*}
\end{prop}
\begin{proof}
Because the number of independent columns of $V$ is greater than or equal to that of $V'$, $N_{\mathrm{orth}}$ is greater than or equal to the latter. Then applying Proposition~\ref{prop: OPA_reduction} to $V', N_{\mathrm{orth}}$ leads to the claim.
\qedsymbol
\end{proof}
Owing to Proposition~\ref{prop: OPA_partial_reduction}, we can orthogonalize the column vectors and drop the zero vectors sequentially during the summation of \Eqref{seq: OPA}. More precisely, the following holds.
\begin{prop}\label{prop: OPA}
We use the notation of Algorithm~\ref{alg: calc_opa}. Let $V$ in \Eqref{seq: OPA} be an input to it and $N_{\mathrm{orth}}$ be an integer greater than the number of linearly independent columns of $V$. If $N_{\mathrm{orth}} \le N_{\mathrm{max}}$, then the summation of \Eqref{seq: OPA} is computed by the output  $V_{\mathrm{orth}}$ of Algorithm \ref{alg: calc_opa}:
\begin{align*}
VV^{\mathrm{T}} =
V_{\mathrm{orth}} V_{\mathrm{orth}}^{\mathrm{T}}.
\end{align*}
\end{prop}
\begin{proof}
This is immediately derived from Proposition~\ref{prop: OPA_partial_reduction}.
\qedsymbol
\end{proof}

Next, we show that $H^+$ can also be written as a summation of factorized forms.
\begin{prop}\label{prop: pseudo_inv_OPA}
Let $V, V_{\mathrm{orth}}$ be the matrices specified in Proposition \ref{prop: OPA}. Then the pseudo-inverse of the GN matrix \Eqref{seq: OPA} can be written in a factorized form and is computed by the output $V_{\mathrm{orth}}^{+}$ of Algorithm \ref{alg: calc_pseudo_opa}:
\begin{align*}
\left(VV^{\mathrm{T}}\right)^{+} =
V_{\mathrm{orth}}^{+} V_{\mathrm{orth}}^{+\mathrm{T}}.
\end{align*}
\end{prop}

\begin{proof}
From the proof of Proposition~\ref{prop: OPA_reduction}, we can write $VV^{\mathrm T}$ as follows:
\begin{align*}
H = VV^{\mathrm{T}}
=
\sum_{i=0}^N \bm{v}_i \bm{v}_i^T
\ \ \ \ 
\bm{v}_i \cdot \bm{v}_j =0 \text{ for } i\ne j
\end{align*}
Hence, if $|\bm{v}_i| > 0$, $\bm{v}_i$ is an eigenvector of $H$ with the positive eigenvalue
$\lambda = \bm{v}_i\cdot \bm{v}_i$.
Let $S$ be the set of vectors of this kind($S = \{v_i|i\in [1,\cdots,N], |\bm{v}_i| > 0\}$).
Because vectors in the orthogonal complement to the subspace spanned by $S$ are eigenvectors of zero eigenvalue, the pseudo-inverse $H^+$ is uniquely determined by the following conditions:
\begin{align*}
H^+ H\bm{w} &= \bm{w}
\text{ if } \bm{w}\in S
\\
H^+ H\bm{w} &=
\bm{0}
\text{ if } \bm{w}\cdot\bm{v}_i = 0
\text{ for } \forall \bm{v}_i\in S
\end{align*}
These can be checked explicitly for
$V_{\mathrm{orth}}^{+}V_{\mathrm{orth}}^{+\mathrm{T}}$
:
\begin{align*}
\left(
V_{\mathrm{orth}}^{+}V_{\mathrm{orth}}^{+\mathrm{T}}
\right)
\left(
VV^{\mathrm{T}}
\right)
\bm{w}
&=
\left(
\sum_{\bm{v} \in S}
\frac{1}{|\bm{v}|^4}\bm{v}\bm{v}^{\mathrm{T}}
\right)
\left(
\sum_{\bm{v} \in S}
\bm{v}\bm{v}^{\mathrm{T}}
\right)
\bm{w}
\\
&=
\begin{cases}
\bm{w}
\text{ if } \bm{w} \in S \\
\bm{0}
\text{ if }
\bm{w}\cdot \bm{v} = 0 \text{ for } \forall \bm{v}\in S
\end{cases}
\end{align*}
Thus, we see that
$H^+ = V_{\mathrm{orth}}^{+}V_{\mathrm{orth}}^{+\mathrm{T}}$.
\qedsymbol
\end{proof}

We have seen that we can implement the accumulation of the vector product elements in \Eqref{seq: OPA} by preparing a buffer, sequentially adding the column vectors from each training task to it, orthogonalizing the elements inside it, and dropping zero elements (Algorithm~\ref{alg: calc_opa}), and that the pseudo-inverse is given by appropriate rescaling of those vectors (Algorithm~\ref{alg: calc_pseudo_opa}).
Note that the number of independent columns in $V$ must be small for the above method to be feasible.
Fortunately, it is expected to be small, and most columns are dropped as zero vectors.
This number corresponds to the non-flat directions of the Hessian, representing the constraints imposed by the loss minimization condition, and typically, it is at most the number of training tasks.

In the experiments of Section~\mref{task_distinction} and \mref{one_step_update} in the main text, for accumulating the vector product elements, $N_{\mathrm{orth}}$ in Algorithm~\ref{alg: calc_opa} is set to 8192 and at each time when the number of accumulated vectors exceeds $N_{\mathrm{max}} = 2 \times N_{\mathrm{orth}}$, the columns are orthogonalized, the vectors with the largest $N_{\mathrm{orth}}$ norms are kept, and the other vectors are dropped. 
After that, we calculate the pseudo-inverse of the Hessian using Algorithm~\ref{alg: calc_pseudo_opa}.
To determine the number of vectors treated as non-zero, we compute the pseudo-inverses under different assumptions about the number of positive eigenvalues and choose the one that minimizes self-rank. We set the assumptions that the Hessian has 64, 128, 256, 512, 1024, 2048, 4096, and 8192 positive eigenvalues. For each of them, we retain the assumed numbers of eigenvalues in descending order, treat the other elements as zero, compute the self-rank using the proposed influence function (\mEqref{eq:influence_ml_gen_param_mod}), and choose the assumption that gives the smallest self-rank. If different assumptions provide the same self-rank, the one with the smallest number of positive eigenvalues is chosen.

\clearpage
\section{Experimental details} \label{experimental_details}
\subsection{Dataset}\label{sup: dataset}
In the experiments below, we use the MiniImagenet and Omniglot datasets. Both are commonly used datasets for measuring model performance in the few-shot learning problem. In addition, to fill their shortcomings, we define the following two datasets.

\paragraph{Synthetic dataset.}
\begin{figure*}[t]
     \centering
     \begin{subfigure}[b]{0.48\linewidth}
         \centering
         \includegraphics[width=\linewidth]
         {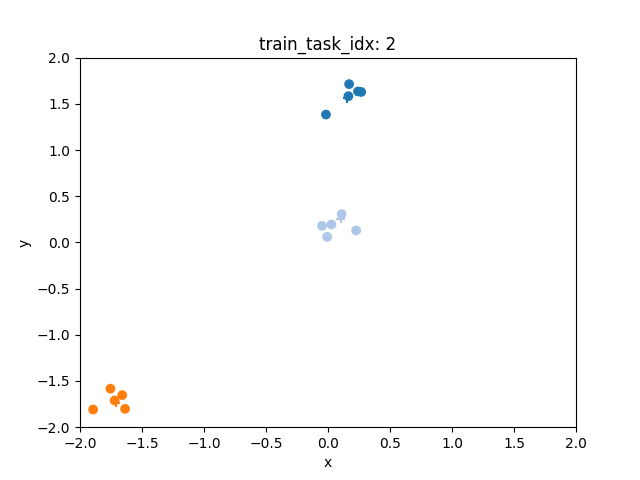}
         \caption{A normal task}
         \label{gc_a}
     \end{subfigure}
     ~
     \begin{subfigure}[b]{0.48\linewidth}
         \centering
         \includegraphics[width=\linewidth]{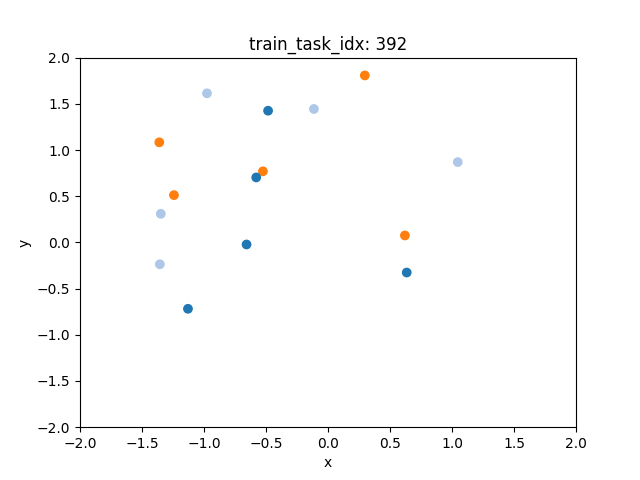}
         \caption{A noise task}
         \label{gc_b}
     \end{subfigure}
    \caption{Examples of synthetic tasks generated from Gaussian distributions. }
    \label{fig: gc_examples}
\end{figure*}
To validate the proposed method with the exact form of \mEqref{eq:influence_ml_gen_param_mod} in the main text, we employ a lightweight network and a taskset that the network can learn easily.
We create training and test tasksets of clustered data points in a two-dimensional space using Gaussian distributions. 
To create each normal task, we first sample cluster centers from a Gaussian distribution with mean 0 (centered at the origin of the plane) and standard deviation 1. 
Then, the members of each cluster are sampled around their center with a standard deviation of 0.1, and a unique label is assigned to each cluster. We also use noise tasks consisting of data points sampled around the origin with a standard deviation of 1 and assigned random labels. 

Fig.~\ref{fig: gc_examples} shows examples of a normal task and a noise task in the 3-way-5-shot problem.

\paragraph{FC60 dataset.}
\begin{table*}[tb]
\centering
\caption{Superclasses and subclasses of FC60 dataset}
\label{tab: train_test_split}
\begin{tabular}{l|l|l}
\toprule
\multicolumn{1}{l}{superclass} & \multicolumn{2}{l}{subclass} \\
{} & train & test \\
\midrule
fish & aquarium fish, flatfish, ray & shark, trout \\
flowers & orchids, poppies, roses & sunflowers, tulips\\
food containers & bottles, bowls, cans & cups, plates\\
fruit and vegetables & apples, mushrooms, oranges & pears, sweet peppers\\
household electrical devices & clock, keyboard, lamp & telephone, television\\
household furniture & bed, chair, couch & table, wardrobe\\
large man-made outdoor things & bridge, castle, house & road, skyscraper\\
large natural outdoor scenes & cloud, forest, mountain & plain, sea\\
reptiles & crocodile, dinosaur, lizard & snake, turtle\\
trees & maple, oak, palm & pine, willow\\
vehicles 1 & bicycle, bus, motorcycle & pickup truck, train\\
vehicles 2 & lawn-mower, rocket, streetcar & tank, tractor\\
\bottomrule
\end{tabular}
\end{table*}
\begin{figure*}[t]
     \centering
     \begin{subfigure}[b]{0.48\linewidth}
         \centering
         \includegraphics[width=\linewidth]
         {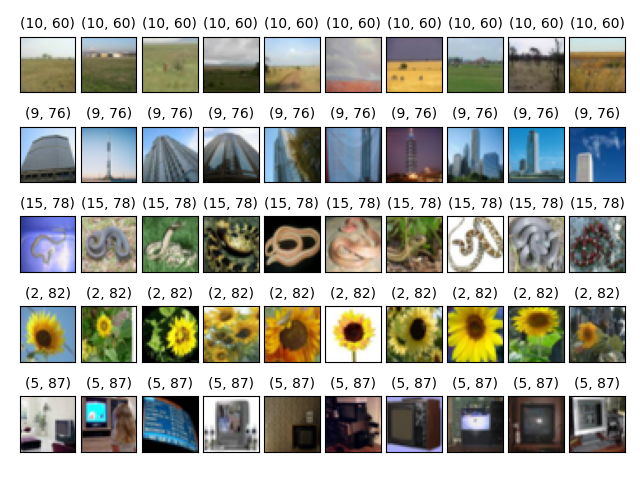}
         \caption{Test task}
         \label{images_a}
     \end{subfigure}
     ~
     \begin{subfigure}[b]{0.48\linewidth}
         \centering
         \includegraphics[width=\linewidth]{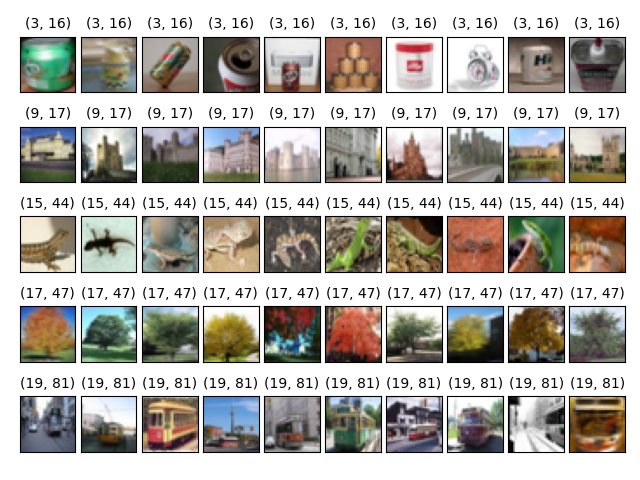}
         \caption{1st training task}
         \label{images_b}
     \end{subfigure}
    \caption{Examples of FC60 tasks. Left: A test task. Right: The training task ranked 1\textsuperscript{st} by TLXML in terms of the influence on the left test task. The pair of labels on each image represents the semantic labels (superclass, subclass) obtained from CIFAR-100. The superclass 9 (large man-made outdoor things) and 15 (reptiles) are shared by them. 
    }
    \label{fig: fc60_examples}
\end{figure*}

To investigate the relation between influence scores and semantic similarity, we need a dataset in which training and test tasks share semantic properties.
We define a new image dataset, \textit{the FC60 dataset}, in which each image is assigned hierarchical labels specifying its super- and sub-classes.
This is achieved by utilizing the FC100 dataset~\citep{oreshkin2018tadam}, which curates few-shot-learning tasks from CIFAR-100~\citep{krizhevsky2009learning}. 
The images of each class in each FC100 are collected from a pair of super- and sub-classes defined in CIFAR100, and the train, validation, and test splits of tasks are constructed in a way that no superclass appears in more than one split.
We split the FC100 training taskset by dividing the subclasses of each of its 12 superclasses into two sets of subclasses for new training and test tasksets.
This results in train/test splits with 60 subclasses in total and 12 superclasses as their parents (see Table~\ref{tab: train_test_split} for details). 
The implementation is based on the FC100 dataset provided by learn2learn and is achieved by applying filters to the subclasses at each data acquisition step.  

Table \ref{tab: train_test_split} shows the superclasses and subclasses used in FC60. Fig.~\ref{fig: fc60_examples} shows examples of a test task and a training task.


\subsection{Distinction of Tasks
} 
\label{sup: task_distinction}
\paragraph{Setup.}
\begin{figure}[t]
\centering
\includegraphics[width=0.6\linewidth]{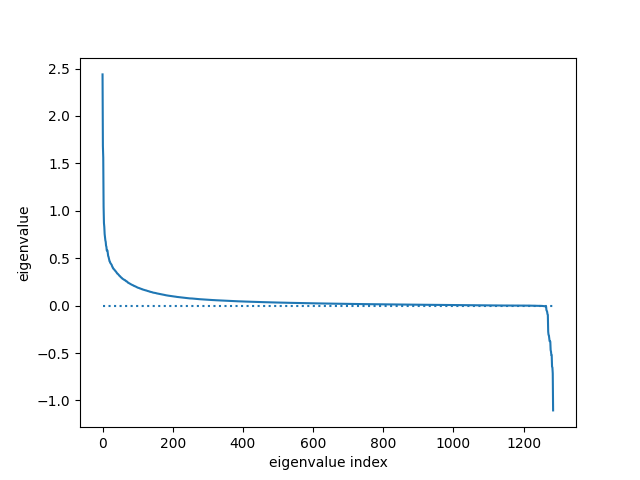}
\caption{Eigenvalues of the Hessian before the pruning in \mref{task_distinction}}
\label{fig: eigenvalue.png}
\end{figure}
We employ MAML as the learning algorithm and train a two-layer fully connected network with widths of 32 and 5, using ReLU activations and batch normalization layers (1,285 parameters) with 1000 meta-batches sampled from 128 training tasks in each dataset. For the MiniImagenet dataset, the model's input is a 32-dimensional feature vector extracted from the image using the Bag-of-Visual-Words~\citep{csurka2002visual} with SIFT descriptors~\citep{lowe1999object,lowe2004distinctive} and k-means clustering. For the Omniglot dataset, the input is a 36-dimensional feature vector extracted from the image by applying 2-dimensional FFT and clipping the 6$\times$6 image at the center. 
\label{pruning_miniimagenet_omniglot}

\begin{table*}[t]
\centering
\caption{Effect of Hessian pruning(MiniImagenet). 
    "\# eigenvalues" denotes the number of eigenvalues 
    treated as positive in computing the pseudo-inverse
    (the 1\textsuperscript{st} row is the original Hessian). 
    The 2\textsuperscript{nd} row corresponds to the pruned Hessian, 
    where all 92 negative eigenvalues are set to zero. 
    The 2\textsuperscript{nd} column shows the self-ranks in the tests without degradation. 
    and the remaining columns show correlation coefficients 
    between the two degradation parameters and the self-ranks or self-scores. 
    Values are reported as means and standard deviations across the 128 tasks.
    $\pm$ means the average and the standard deviation across 128 test tasks.
    }
\label{tab: pruning_mi}
\begin{tabular}{cccccc}\hline
\multicolumn{1}{c}{\# eigenvalues} &
\multicolumn{1}{c}{self\-rank(avg$\pm$std)} &
\multicolumn{4}{c}{correlation with degradation(avg$\pm$std)} \\
 & & alpha/rank & alpha/score & ratio/rank & ratio/score \\
\hline
1285 & 12.6$\pm$18.9 & 0.51$\pm$0.32 & -0.41$\pm$0.29 & 0.36$\pm$0.32 & -0.11$\pm$0.31\\
1193 & 0.0$\pm$0.0 & 0.69$\pm$0.21 & -0.69$\pm$0.22 & 0.46$\pm$0.30 & -0.15$\pm$0.34\\
512 & 0.0$\pm$0.0 & 0.71$\pm$0.12 & -0.96$\pm$0.06 & 0.63$\pm$0.09 & -0.89$\pm$0.13\\
256 & 0.0$\pm$0.0 & 0.71$\pm$0.11 & -0.95$\pm$0.08 & 0.62$\pm$0.11 & -0.88$\pm$0.14\\
128 & 0.0$\pm$0.0 & 0.72$\pm$0.10 & -0.94$\pm$0.04 & 0.63$\pm$0.10 & -0.86$\pm$0.15\\
64 & 0.0$\pm$0.0 & 0.71$\pm$0.12 & -0.92$\pm$0.06 & 0.66$\pm$0.12 & -0.77$\pm$0.20\\
32 & 0.0$\pm$0.2 & 0.72$\pm$0.16 & -0.85$\pm$0.12 & 0.68$\pm$0.14 & -0.68$\pm$0.22\\
16 & 2.0$\pm$3.2 & 0.67$\pm$0.22 & -0.69$\pm$0.20 & 0.61$\pm$0.22 & -0.46$\pm$0.27\\
8 & 8.6 $\pm$9.1 & 0.55$\pm$0.26 & -0.53$\pm$0.23 & 0.47$\pm$0.27 & -0.29$\pm$0.31\\
\hline
\end{tabular}
\end{table*}
\begin{table*}[t]
\centering
\caption{Effect of pruning the Hessian(Omniglot). 
See table~\ref{tab: pruning_mi} for notations.
The second row is the case that we only prune the 428 negative eigenvalues.
There are cases in which increasing $\alpha$ does not change the self-ranks. The numbers of those cases are shown in the brackets, and they are removed from the statistics because the correlation coefficients are not defined for them.}
\label{table: pruning_omniglot}

\begin{tabular}{cccccc}\hline
\multicolumn{1}{c}{\# eigenvalues} &
\multicolumn{1}{c}{self\-rank(avg$\pm$std)} &
\multicolumn{4}{c}{correlation with degradation(avg$\pm$std)} \\
 & & alpha/rank & alpha/score & ratio/rank & ratio/score \\
\hline
1413&66.0 $\pm$36.9&0.15 $\pm$0.48(21)&0.06 $\pm$0.50&0.02 $\pm$0.34&0.03 $\pm$0.23 \\
985&7.8 $\pm$10.0&0.48 $\pm$0.12(2)&-0.37 $\pm$0.33&0.25 $\pm$0.28&0.01 $\pm$0.23 \\
512&0.8 $\pm$5.0&0.50 $\pm$0.00(1)&-0.50 $\pm$0.00&0.35 $\pm$0.26&-0.15 $\pm$0.23 \\
256&1.6 $\pm$6.3&0.50 $\pm$0.00(1)&-0.50 $\pm$0.00&0.34 $\pm$0.27&-0.12 $\pm$0.25 \\
128&2.9 $\pm$7.4&0.50 $\pm$0.00(1)&-0.50 $\pm$0.00&0.36 $\pm$0.27&-0.11 $\pm$0.24 \\
64&4.3 $\pm$8.0&0.50 $\pm$0.00(1)&-0.50 $\pm$0.00&0.36 $\pm$0.26&-0.13 $\pm$0.23 \\
32&6.4 $\pm$8.8&0.50 $\pm$0.00(1)&-0.49 $\pm$0.09&0.33 $\pm$0.26&-0.08 $\pm$0.24 \\
16&8.4 $\pm$10.0&0.50 $\pm$0.00(1)&-0.50 $\pm$0.00&0.32 $\pm$0.30&-0.08 $\pm$0.25 \\
8&11.8 $\pm$12.5&0.50 $\pm$0.00(4)&-0.50 $\pm$0.00&0.29 $\pm$0.29&-0.06 $\pm$0.24 \\
\hline
\end{tabular}
\end{table*}
\begin{figure*}[tb]
     \centering
     \begin{subfigure}[b]{0.48\linewidth}
         \centering
         \includegraphics[width=\linewidth]{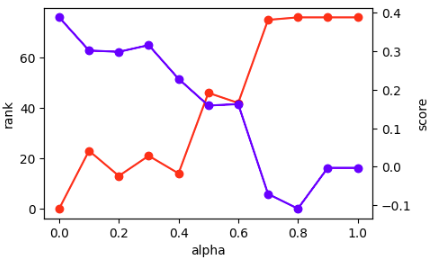}
         \caption{Example of the effect of increasing $\alpha$. 
         $\alpha=0$ means the test task without degradation.
         $\alpha{=}1$ means all the images in the test task are completely black.}
     \end{subfigure}
     ~
     \begin{subfigure}[b]{0.48\linewidth}
         \centering
         \includegraphics[width=\linewidth]{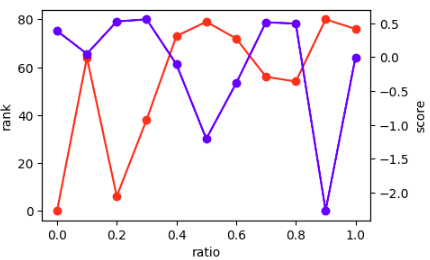}
         \caption{Example of the effect of increasing the ratio. 
         ratio=0 means the test task without degradation.
         ratio=1 means all the images in the test task are dark.}
     \end{subfigure}
    \caption{Test with degraded training tasks(MiniImagenet). 
    The parameters $\alpha$ and \textit{ratio} specify the darkness of images, 
    and the number of dark images in each degraded task, respectively. The red and blue lines represent ranks and scores, respectively. Both examples were performed with the Hessian pruned to retain only the 1193 most significant eigenvalues. }
    \label{test-with-degraded-tasks}
\end{figure*}

\paragraph{MiniImagenet}
In the experiments in Section~\mref{task_distinction} with MiniImagenet, we encounter negative eigenvalues of the Hessian. We provide some details here. Figure~\ref{fig: eigenvalue.png} shows 1285 eigenvalues arranged from the largest to the smallest. We observe that large positive eigenvalues mainly appear in the first several hundred elements, and also that negative eigenvalues appear in the tail. 

In addition to the results described in the main paper, as a sanity check, we checked that reducing the similarity between training and test tasks worsens the self-ranks. 
We degrade the test tasks by darkening their images and examine whether the ranks and scores of the originally identical training tasks (referred to as \textit{self-ranks} and \textit{self-scores}) worsen as the similarity decreases. 

Figure~\ref{test-with-degraded-tasks} shows examples of the results obtained with MiniImagenet.
We observe that the ranks and scores tend to get worse
as the darkness (parameterized by $\alpha$) or the portion (parametrized by \textit{ratio}) of dark images in the test task increases.  
Those are example plots with a small amount of Hessian pruning.  
We investigate the effects of pruning on the adequacy of the influence scores by looking at the correlations between the degradation parameters ($\alpha$ and ratio) and the two measures(self-ranks and self-scores), which are shown in the third to sixth columns of Table~\ref{tab: pruning_mi}.
We observe that appropriate pruning improves those correlation values.

\paragraph{Omniglot}
We also conduct experiments with the Omniglot dataset. 
In this case, we encounter 428 negative eigenvalues of the Hessian. 
The effects of pruning on the self-rank are shown in the first column of Table \ref{table: pruning_omniglot}. Tests with the degradation of those test tasks are also conducted. 
Again, we observe that the correlations improve to some extent by pruning. However, those correlations are weaker than in the cases of MiniImagenet (Table \ref{tab: pruning_mi}). 
A possible reason is that the tasks in Omniglot are easier to adapt to than those in MiniImagenet, which makes the trained model less sensitive to degradation.

\clearpage

\subsection{Distinction of Normal and Noise Task Distributions
}
\label{sup: noise_distinction}
\begin{table*}[t]
\centering
\caption{
Experiments with a CNN trained by MAML(MiniImagenet dataset, combined with noise image tasks).
128 test tasks were selected from the test taskset of pure MiniImagenet. 
The numbers of proper tests are shown in the second-to-last column for each training setting. The standard deviation $\sigma$ under the null-hypothesis of random ordering is $\sqrt{128\times0.5^2} \sim 5.66$. $\pm$ means the average and standard deviation across 5 runs of training.
}
\label{tab: noise_distinction_maml_mi}
\begin{tabular}{cccccccc}
\toprule
Meta-learner & Dataset & \multicolumn{2}{l}{Training tasks} & \multicolumn{2}{l}{Accuracy} & \multicolumn{2}{l}{Proper tests} \\
             {} &      {} &          total & noise &            test &           train &       [count] &            [$\sigma$] \\
\midrule
           MAML &      MI &            128 &    16 & 0.314$\pm$0.003 & 1.000$\pm$0.000 & 16.4$\pm$12.9 & -8.4$\pm$2.3 $\sigma$ \\
           MAML &      MI &            256 &    32 & 0.323$\pm$0.014 & 1.000$\pm$0.000 &  14.4$\pm$7.6 & -8.8$\pm$1.3 $\sigma$ \\
           MAML &      MI &            512 &    64 & 0.345$\pm$0.006 & 0.995$\pm$0.005 & 51.2$\pm$16.5 & -2.3$\pm$2.9 $\sigma$ \\
           MAML &      MI &           1024 &   128 & 0.368$\pm$0.007 & 0.902$\pm$0.049 & 99.6$\pm$20.5 &  6.3$\pm$3.6 $\sigma$ \\
           MAML &      MI &           2048 &   256 & 0.435$\pm$0.006 & 0.828$\pm$0.014 &  73.4$\pm$6.0 &  1.7$\pm$1.1 $\sigma$ \\
           MAML &      MI &           4096 &   512 & 0.520$\pm$0.012 & 0.762$\pm$0.008 & 89.2$\pm$10.3 &  4.5$\pm$1.8 $\sigma$ \\
           MAML &      MI &           8192 &  1024 & 0.552$\pm$0.012 & 0.717$\pm$0.007 & 85.6$\pm$11.1 &  3.8$\pm$2.0 $\sigma$ \\
\bottomrule
\end{tabular}
\end{table*}

\paragraph{Setup.}
For the networks used in the experiments below, we employ ReLu activation and batch normalization.

\paragraph{Synthetic dataset.}
To examine the method with the exact form of 
\mEqref{eq:influence_ml_gen_params}, we employ a lightweight network and easy tasksets that can be learned easily.
Using 1024 training tasks of the 3-ways-5-shots problem, including 128 noise tasks, in the synthetic dataset described in \ref{sup: dataset}, we train a 3-layer fully connected network with 2-dimensional input, 4-4 hidden dimensions, 3-dimensional output layers (63 parameters) with 30,000 meta-batches, and then calculated $I^{\rm{meta}}$. 
To determine the number of positive eigenvalues of an exact Hessian, average self-ranks across the training tasks are calculated under different assumptions about the number of positive eigenvalues (4, 8, 16, 32, and 63). 
32 is chosen as the one giving the best average self-rank.
The result is described in the main paper.

Additionally, we calculate the scores using an approximated Hessian. To obtain the Hessian, we accumulate vector products across the training tasks using a buffer for 32 factorized elements. 
\footnote{In terms of the parameters in Algorithm~\ref{alg: calc_opa}, $N_{\mathrm{orth}}=32, N_{\mathrm{max}}=2\times 32$},
We obtain a result similar to that for the exact Hessian and observe that 124 of 128 tests yield the proper order of the training task distributions, with their positions defined by mean values. This count exceeds the average count ($=64$) of the binomial distribution by $10.6 \sigma$ under the hypothesis of random ordering. We also check the correlations between the scores calculated with the exact and approximated Hessians. The Pearson correlation of the influence scores for the training tasks calculated for each of the 128 tests is 0.715$\pm$0.092.

\paragraph{MiniImagenet and Omniglot.}
\begin{figure}[th]
    \centering
    \includegraphics[width=0.8\textwidth]
    {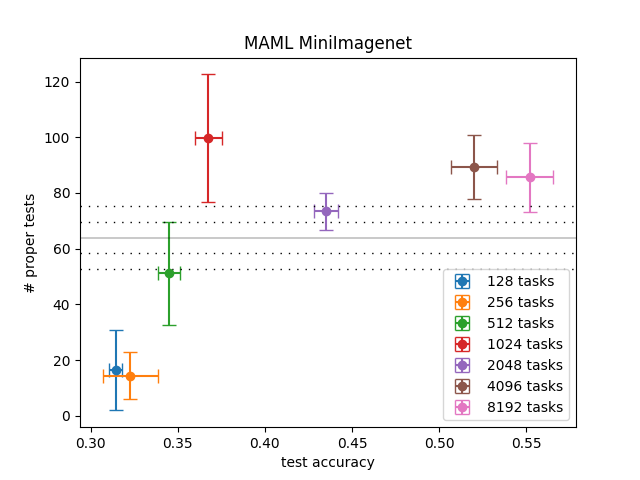}
    \caption{Effect of generalization with a CNN trained by MAML(MiniImagenet dataset, combined with noise image tasks).
    The solid horizontal line at y=64 represents the mean value under the hypothesis of random ordering. The dashed horizontal lines represent the mean value $\pm1\sigma$ and $\pm2\sigma$ under the hypothesis of random ordering. 
   }
    \label{fig: noise_distinction_maml}
\end{figure}
We train 3-conv$\times$32-filter + 1-fully-connected layers networks (21,029 parameters for MiniImagenet, 19,178 parameters for Omniglot) using MAML and 4-conv $\times$32-filter feature extractors (28,896 parameters for MiniImagenet, 28,320 parameters for Omniglot) using Protonet
\footnote{
We choose networks such that the number of model parameters is around $\sim$20,000, and the test accuracy does not decrease clearly in the tails of the learning curves during the training with 8192 tasks from the dataset in each setting.
}.
We use 128, 256, 512, 1024, 2048, 4096, and 8192 training tasks for each dataset, and combine them with 16, 32, 64, 128, 256, 512, and 1024 noise tasks, respectively.
We create noise tasks by replacing the image tensors in each training task with uniform noise tensors of the same shape(noise images).
In MAML training, the networks are trained with 80,000 meta-batches for MiniImagenet and 40,000 meta-batches for Omniglot.
In Protonet training, the networks are trained with 20,000 meta-batches for MiniImagenet and 10,000 meta-batches for Omniglot.
We conduct 5 training runs with seeds 0, 1, 2, 3, and 42.
We evaluate the influence scores of the training tasks on 128 test tasks, and count the number of proper tests based on \mEqref{eq:influence_ml_gen_adpt}, and \mEqref{eq:influence_ml_gen_perf} with the projected influence on the meta-parameters (\mEqref{eq:influence_ml_gen_param_mod}) and the Gauss-Newton matrix approximation of the Hessian (\mEqref{eq: OPA}).
When we calculate an approximated Hessian (\mEqref{eq: OPA}) after each training, we accumulate vector products over the training tasks with the buffer size set to 8192
\footnote{In terms of the parameters in Algorithm~\ref{alg: calc_opa}, $N_{\mathrm{orth}}=8192, N_{\mathrm{max}}=2\times 8192$},
and then truncate them before evaluating the influence scores. To determine the truncated number, we compute the average self-rank across the training tasks under different assumptions about the number of positive eigenvalues (64, 128, 256, 512, 1024, 2048, 4096, and 8192). 
We choose the most appropriate assumption that gives the best self-rank. When different assumptions provide the same self-rank (like $0\pm0$), we choose the assumption of the smallest number.

Table
\ref{tab: noise_distinction_maml_mi}, 
\ref{tab: noise_distinction_maml_om}, 
\ref{tab: noise_distinction_protonet_mi}, and
\ref{tab: noise_distinction_protonet_om}
show the results of the experiments.
We observe that when the number of training tasks is sufficiently large, the number of proper tests negates the hypothesis of random ordering, suggesting that our scoring method satisfies Property 2 statistically. 
Noting that the statistical significance is weak in the results of a small number of training tasks, we can see that Property 2 arises as an effect of generalization. Fig.\ref{fig: noise_distinction_maml} demonstrates this as the relation between the test accuracy and the number of proper tests in the case of MAML and MiniImagenet. In that case, when the training taskset is small (128 and 256 training tasks), the number of proper tests is exceeded by the number of not-proper tests. This can be interpreted as reflecting the tendency that, when overfitting occurs, only a few training tasks similar to the test task provide useful information, and most regular training tasks become detrimental.
\begin{table*}[t]
\centering
\caption{
Experiments with a CNN trained by MAML(Omniglot dataset, combined with noise image tasks). 
See Table \ref{tab: noise_distinction_maml_mi} for notations.
}
\label{tab: noise_distinction_maml_om}
\begin{tabular}{cccccccc}
\toprule
Meta-learner & Dataset & \multicolumn{2}{l}{Training tasks} & \multicolumn{2}{l}{Accuracy} & \multicolumn{2}{l}{Proper tests} \\
             {} &      {} &          total & noise &            test &           train &       [count] &           [$\sigma$] \\
\midrule
           MAML &      OM &            128 &    16 & 0.668$\pm$0.017 & 1.000$\pm$0.000 &  68.6$\pm$6.0 & 0.8$\pm$1.1 $\sigma$ \\
           MAML &      OM &            256 &    32 & 0.759$\pm$0.009 & 1.000$\pm$0.000 &  77.0$\pm$7.9 & 2.3$\pm$1.4 $\sigma$ \\
           MAML &      OM &            512 &    64 & 0.824$\pm$0.010 & 0.984$\pm$0.032 & 118.4$\pm$4.4 & 9.6$\pm$0.8 $\sigma$ \\
           MAML &      OM &           1024 &   128 & 0.882$\pm$0.032 & 0.999$\pm$0.003 & 114.8$\pm$7.2 & 9.0$\pm$1.3 $\sigma$ \\
           MAML &      OM &           2048 &   256 & 0.892$\pm$0.005 & 0.987$\pm$0.004 &  90.0$\pm$3.9 & 4.6$\pm$0.7 $\sigma$ \\
           MAML &      OM &           4096 &   512 & 0.933$\pm$0.006 & 0.940$\pm$0.005 &  82.2$\pm$5.9 & 3.2$\pm$1.0 $\sigma$ \\
           MAML &      OM &           8192 &  1024 & 0.960$\pm$0.001 & 0.906$\pm$0.003 &  94.0$\pm$6.0 & 5.3$\pm$1.1 $\sigma$ \\
\bottomrule
\end{tabular}

\end{table*}
\begin{table*}[t]
\centering
\caption{
Experiments with a CNN trained by Prototypical Network (MiniImagenet dataset, combined with noise image tasks). 
See Table \ref{tab: noise_distinction_maml_mi} for notations.
}
\label{tab: noise_distinction_protonet_mi}
\begin{tabular}{cccccccc}
\toprule
Meta-learner & Dataset & \multicolumn{2}{l}{Training tasks} & \multicolumn{2}{l}{Accuracy} & \multicolumn{2}{l}{Proper tests} \\
             {} &      {} &          total & noise &            test &           train &       [count] &            [$\sigma$] \\
\midrule
       Protonet &      MI &            128 &    16 & 0.397$\pm$0.016 & 0.997$\pm$0.001 & 63.0$\pm$15.5 & -0.2$\pm$2.7 $\sigma$ \\
       Protonet &      MI &            256 &    32 & 0.435$\pm$0.018 & 0.778$\pm$0.010 & 108.2$\pm$9.2 &  7.8$\pm$1.6 $\sigma$ \\
       Protonet &      MI &            512 &    64 & 0.496$\pm$0.009 & 0.608$\pm$0.037 &  94.4$\pm$8.4 &  5.4$\pm$1.5 $\sigma$ \\
       Protonet &      MI &           1024 &   128 & 0.503$\pm$0.015 & 0.570$\pm$0.007 &  81.8$\pm$7.5 &  3.1$\pm$1.3 $\sigma$ \\
       Protonet &      MI &           2048 &   256 & 0.498$\pm$0.007 & 0.550$\pm$0.008 &  84.2$\pm$7.6 &  3.6$\pm$1.3 $\sigma$ \\
       Protonet &      MI &           4096 &   512 & 0.497$\pm$0.016 & 0.537$\pm$0.007 &  96.4$\pm$9.9 &  5.7$\pm$1.7 $\sigma$ \\
       Protonet &      MI &           8192 &  1024 & 0.509$\pm$0.007 & 0.535$\pm$0.008 &  95.6$\pm$5.3 &  5.6$\pm$0.9 $\sigma$ \\
\bottomrule
\end{tabular}
\end{table*}
\begin{table*}[t]
\centering
\caption{
Experiments with a CNN trained by Prototypical Network(Omniglot dataset, combined with noise image tasks). 
See Table \ref{tab: noise_distinction_maml_mi} for notations.
}
\label{tab: noise_distinction_protonet_om}
\begin{tabular}{cccccccc}
\toprule
Meta-learner & Dataset & \multicolumn{2}{l}{Training tasks} & \multicolumn{2}{l}{Accuracy} & \multicolumn{2}{l}{Proper tests} \\
             {} &      {} &          total & noise &            test &           train &       [count] &            [$\sigma$] \\
\midrule
       Protonet &      OM &            128 &    16 & 0.937$\pm$0.007 & 1.000$\pm$0.000 &  77.4$\pm$4.7 &  2.4$\pm$0.8 $\sigma$ \\
       Protonet &      OM &            256 &    32 & 0.958$\pm$0.003 & 1.000$\pm$0.000 &  61.0$\pm$3.4 & -0.5$\pm$0.6 $\sigma$ \\
       Protonet &      OM &            512 &    64 & 0.963$\pm$0.004 & 0.986$\pm$0.004 & 112.6$\pm$3.9 &  8.6$\pm$0.7 $\sigma$ \\
       Protonet &      OM &           1024 &   128 & 0.973$\pm$0.002 & 0.928$\pm$0.002 &  84.0$\pm$4.1 &  3.5$\pm$0.7 $\sigma$ \\
       Protonet &      OM &           2048 &   256 & 0.978$\pm$0.002 & 0.901$\pm$0.001 &  81.0$\pm$4.1 &  3.0$\pm$0.7 $\sigma$ \\
       Protonet &      OM &           4096 &   512 & 0.978$\pm$0.003 & 0.891$\pm$0.001 &  84.0$\pm$4.6 &  3.5$\pm$0.8 $\sigma$ \\
       Protonet &      OM &           8192 &  1024 & 0.979$\pm$0.001 & 0.884$\pm$0.008 &  81.2$\pm$5.5 &  3.0$\pm$1.0 $\sigma$ \\
\bottomrule
\end{tabular}
\end{table*}

\subsection{Consistency with Semantic Similarity}
\label{sup: label_overlap}
\begin{table*}[t]
\centering
\caption{Experiments of distinguishing a training task subpopulation with superclasses shared with the test task from other training tasks. The 3rd column specifies the least number of superclasses shared with each test task, which defines the training tasks 'similar' to the test task. The 5th column shows the number of training tasks satisfying the condition. The 7th column shows the counts of proper tests and their statistical significance. $\pm$ means the average and standard deviations across 1024 test tasks.
Expl. time means the time for generating explanation data, that is, the time spent for computing the 1024 test losses and the influence scores of the 8192 training tasks for each of them with a given $I^{\rm meta}$ for TLXML and the time spent for fine-tuning with the 1024 test tasks, computing the task embedding of each test task and the similarities with given 8192 training task embeddings for Task2Vec.
}
\label{tab: label_overlap}

\begin{tabular}{cccccccc}
\toprule
Method &   Expl. time &     Filter & \multicolumn{2}{l}{Train tasks} & \multicolumn{2}{l}{Test tasks} & $\sigma$ \\
{} & {} & label overlap &       total &     filtered &      total &              proper &       {} \\
\midrule
           MAML+TLXML & 33sec            & 1 &        8192 & 7522$\pm$314 &       1024 &  618 (+6.6$\sigma$) &       16 \\
           MAML+TLXML & 33sec            & 2 &        8192 & 4781$\pm$718 &       1024 & 676 (+10.3$\sigma$) &       16 \\
           MAML+TLXML & 33sec            & 3 &        8192 & 1508$\pm$422 &       1024 & 688 (+11.0$\sigma$) &       16 \\
           MAML+TLXML & 33sec            & 4 &        8192 &   171$\pm$69 &       1024 &  637 (+7.8$\sigma$) &       16 \\
\midrule
       Protonet+TLXML & 56 sec            & 1 &        8192 & 7522$\pm$314 &       1024 &  658 (+9.1$\sigma$) &       16 \\
       Protonet+TLXML & 56 sec            & 2 &        8192 & 4781$\pm$718 &       1024 & 672 (+10.0$\sigma$) &       16 \\
       Protonet+TLXML & 56 sec            & 3 &        8192 & 1508$\pm$422 &       1024 & 703 (+11.9$\sigma$) &       16 \\
       Protonet+TLXML & 56 sec            & 4 &        8192 &   171$\pm$69 &       1024 &  644 (+8.3$\sigma$) &       16 \\
\midrule
       Task2Vec(8 steps) & 100sec &   1 & 8192 & 7522$\pm$314  &  1024 & 554 (+2.6$\sigma$) & 16\\
       Task2Vec(8 steps)& 100sec &   2 & 8192 & 4781$\pm$718  & 1024 & 574 (+3.9$\sigma$) & 16 \\
       Task2Vec(8 steps)& 100sec &   3 & 8192 & 1508$\pm$422  & 1024 & 568 (+3.5$\sigma$)  & 16 \\
       Task2Vec(8 steps)& 100sec &   4 & 8192 & 171$\pm$69  & 1024 & 574 (+3.9$\sigma$) & 16 \\
\midrule
       Task2Vec(16 steps) & 107sec &   1 & 8192 & 7522$\pm$314  &  1024 & 551 (+2.4$\sigma$) & 16\\
       Task2Vec(16 steps)& 107sec &   2 & 8192 & 4781$\pm$718  & 1024 & 583 (+4.4$\sigma$)  & 16 \\
       Task2Vec(16 steps)& 107sec &   3 & 8192 & 1508$\pm$422  & 1024 & 572 (+3.8$\sigma$) & 16 \\
       Task2Vec(16 steps)& 107sec &   4 & 8192 & 171$\pm$69  & 1024 & 541 (+1.8$\sigma$) & 16 \\
\midrule
       Task2Vec(32 steps) & 124sec &   1 & 8192 & 7522$\pm$314  &  1024 & 599 (+5.4$\sigma$) & 16\\
       Task2Vec(32 steps)& 124sec &   2 & 8192 & 4781$\pm$718  & 1024 & 599 (+5.4$\sigma$) & 16 \\
       Task2Vec(32 steps)& 124sec &   3 & 8192 & 1508$\pm$422  & 1024 & 599 (+5.4$\sigma$) & 16 \\
       Task2Vec(32 steps)& 124sec &   4 & 8192 & 171$\pm$69  & 1024 & 586 (+4.6$\sigma$) & 16 \\
\midrule
       Task2Vec(64 steps) & 155sec &   1 & 8192 & 7522$\pm$314  &  1024 & 653 (+8.8$\sigma$) & 16 \\
       Task2Vec(64 steps)& 155sec &   2 & 8192 & 4781$\pm$718  & 1024 & 657 (+9.1$\sigma$) & 16 \\
       Task2Vec(64 steps) & 155sec &   3 & 8192 & 1508$\pm$422  &  1024 & 665 (+9.6$\sigma$)  & 16 \\
       Task2Vec(64 steps)& 155sec &   4 & 8192 & 171$\pm$69  & 1024 & 654 (+8.9$\sigma$) & 16 \\
\bottomrule
\end{tabular}
\end{table*}
\begin{table*}[t]
\centering
\caption{Experiments of distinguishing a training task subpopulation with superclasses shared with the test task (classified by recall values) from other training tasks (MAML). The 2nd column specifies the recall value of each label in the test tasks, and the 3rd column specifies the least number of labels with that recall value. The 6th column shows the number of test tasks whose labels satisfy the conditions. The 5th column shows the number of training tasks 'similar' to the test task, meaning that the superlabels in each of them include those used for conditioning the test task. The 7th column shows the counts of proper tests and their statistical significance. $\pm$ means the average and standard deviations across the test tasks specified by the conditions.}
\label{tab: label_overlap_recall_maml}
\begin{tabular}{cccccccc}
\toprule
Method & \multicolumn{2}{l}{Filter} & \multicolumn{2}{l}{Train tasks} & \multicolumn{2}{l}{Test tasks} & $\sigma$ \\
             {} & recall & label overlap &       total &      filtered &      total &             proper &       {} \\
\midrule
           MAML+TLXML &    1.0 &             1 &        8192 & 2013$\pm$1145 &        828 & 440 (+1.8$\sigma$) &     14.4 \\
           MAML+TLXML &    1.0 &             2 &        8192 &   785$\pm$325 &        377 & 205 (+1.7$\sigma$) &      9.7 \\
           MAML+TLXML &    1.0 &             3 &        8192 &    245$\pm$53 &         97 &  57 (+1.7$\sigma$) &      4.9 \\
\midrule
           MAML+TLXML &    0.8 &             1 &        8192 & 1833$\pm$1184 &        819 & 488 (+5.5$\sigma$) &     14.3 \\
           MAML+TLXML &    0.8 &             2 &        8192 &   728$\pm$358 &        427 & 255 (+4.0$\sigma$) &     10.3 \\
           MAML+TLXML &    0.8 &             3 &        8192 &    221$\pm$81 &        136 &  90 (+3.8$\sigma$) &      5.8 \\
           MAML+TLXML &    0.8 &             4 &        8192 &     43$\pm$17 &         20 &  17 (+3.1$\sigma$) &      2.2 \\
 \midrule
           MAML+TLXML &    0.6 &             1 &        8192 & 2187$\pm$1105 &        706 & 431 (+5.9$\sigma$) &     13.3 \\
           MAML+TLXML &    0.6 &             2 &        8192 &   793$\pm$319 &        267 & 170 (+4.5$\sigma$) &      8.2 \\
           MAML+TLXML &    0.6 &             3 &        8192 &    217$\pm$81 &         55 &  37 (+2.6$\sigma$) &      3.7 \\
\midrule
           MAML+TLXML &    0.4 &             1 &        8192 &  2485$\pm$960 &        549 & 319 (+3.8$\sigma$) &     11.7 \\
           MAML+TLXML &    0.4 &             2 &        8192 &   266$\pm$848 &        138 &  87 (+3.1$\sigma$) &      5.9 \\
           MAML+TLXML &    0.4 &             3 &        8192 &    247$\pm$41 &         16 &  11 (+1.5$\sigma$) &      2.0 \\
\midrule
           MAML+TLXML &    0.2 &             1 &        8192 &  2707$\pm$764 &        380 & 223 (+3.4$\sigma$) &      9.7 \\
           MAML+TLXML &    0.2 &             2 &        8192 &   906$\pm$226 &         57 &  35 (+1.7$\sigma$) &      3.8 \\
\midrule           
           MAML+TLXML &    0.0 &             1 &        8192 &   2888$\pm$87 &        220 & 133 (+3.1$\sigma$) &      7.4 \\
\bottomrule
\end{tabular}
\end{table*}

\begin{table*}[t]
\centering
\caption{Experiments of distinguishing a training task subpopulation with superclasses shared with the test task (classified by recall values) from other training tasks (Protonet). See Table \ref{tab: label_overlap_recall_maml} for notations}
\label{tab: label_overlap_recall_protonet}
\begin{tabular}{cccccccc}
\toprule
Method & \multicolumn{2}{l}{Filter} & \multicolumn{2}{l}{Train tasks} & \multicolumn{2}{l}{Test tasks} & $\sigma$ \\
             {} & recall & label overlap &       total &      filtered &      total &             proper &       {} \\
\midrule
       Protonet+TLXML &    1.0 &             1 &        8192 & 2225$\pm$1095 &        681 & 396 (+4.3$\sigma$) &     13.0 \\
       Protonet+TLXML &    1.0 &             2 &        8192 &   804$\pm$326 &        248 & 151 (+3.4$\sigma$) &      7.9 \\
       Protonet+TLXML &    1.0 &             3 &        8192 &    218$\pm$82 &         56 &  37 (+2.4$\sigma$) &      3.7 \\
\midrule
       Protonet+TLXML &    0.8 &             1 &        8192 & 1652$\pm$1197 &        838 & 479 (+4.1$\sigma$) &     14.5 \\
       Protonet+TLXML &    0.8 &             2 &        8192 &   679$\pm$378 &        492 & 290 (+4.0$\sigma$) &     11.1 \\
       Protonet+TLXML &    0.8 &             3 &        8192 &    207$\pm$90 &        184 & 109 (+2.5$\sigma$) &      6.8 \\
       Protonet+TLXML &    0.8 &             4 &        8192 &     45$\pm$15 &         36 &  25 (+2.3$\sigma$) &      3.0 \\
 \midrule
       Protonet+TLXML &    0.6 &             1 &        8192 & 1874$\pm$1198 &        790 & 478 (+5.9$\sigma$) &     14.1 \\
       Protonet+TLXML &    0.6 &             2 &        8192 &   697$\pm$373 &        392 & 233 (+3.7$\sigma$) &      9.9 \\
       Protonet+TLXML &    0.6 &             3 &        8192 &    194$\pm$92 &        127 &  68 (+0.8$\sigma$) &      5.6 \\
       Protonet+TLXML &    0.6 &             4 &        8192 &     33$\pm$18 &         20 &  12 (+0.9$\sigma$) &      2.2 \\
 \midrule
       Protonet+TLXML &    0.4 &             1 &        8192 & 2311$\pm$1075 &        589 & 341 (+3.8$\sigma$) &     12.1 \\
       Protonet+TLXML &    0.4 &             2 &        8192 &   763$\pm$344 &        187 & 118 (+3.6$\sigma$) &      6.8 \\
       Protonet+TLXML &    0.4 &             3 &        8192 &    162$\pm$88 &         37 &  21 (+0.8$\sigma$) &      3.0 \\
 \midrule
       Protonet+TLXML &    0.2 &             1 &        8192 &  2763$\pm$718 &        312 & 174 (+2.0$\sigma$) &      8.8 \\
       Protonet+TLXML &    0.2 &             2 &        8192 &   855$\pm$244 &         38 &  19 (+0.0$\sigma$) &      3.1 \\
 \midrule
       Protonet+TLXML &    0.0 &             1 &        8192 &  2999$\pm$243 &         74 &  44 (+1.6$\sigma$) &      4.3 \\
\bottomrule
\end{tabular}
\end{table*}
Using 8192 training tasks in the FC60 dataset, which is described \ref{sup: dataset}, we train 3-conv$\times$16-filter + 1-fully-connected layers networks (6469 parameters) with 40,000 meta-batches using MAML and 4-conv$\times$32-filter feature extractors (28,896 parameters) with 40,000 meta-batches using Protonet. 

Table \ref{tab: label_overlap} presents the complete list of results from counting the proper tests based on the number of superclasses shared between the training and test tasks. This demonstrates that TLXML effectively identifies subpopulations in the training taskset that share semantic properties with the test tasks, distinguishing them from other training tasks.

Since TLXML enables us to quantify the influence of training tasks on various aspects of adaptation and inference through the chain rule of differentiation, it is hard to define a baseline with a comparable range of functionality. However, we can set a baseline for each role it can play.
As a baseline for the current experimental setting, we use Task2Vec~\citep{achille2019task2vec}, one of the most established methods for measuring task similarity. 
Task2Vec computes task embeddings using the Fisher information of a network obtained by pre-training a shared base network and fine-tuning it for each task. It also provides a measure of task similarity, computed as the cosine similarity between the embeddings of two tasks. 
To use it as a baseline of the experiment, we take the following path:
\begin{itemize}
\item Network: We use the same network structure as in the MAML experiment for fair comparison.
\item Pre-training: We replace the network head with a linear layer with an output size of 100 and train it with CIFAR-100 for 100 epochs using Adam with a batch size of 128 and a learning rate of 0.001. CIFAR-100 is a reasonable choice because FC60 is based on its image data.
\item Embedding training tasks: Using 8192 FC60 training tasks in the 5-way-5-shot problem setting, we fine-tune the network head with Adam. For each task, the head is initialized, and all 50 data points are used in each Adam update. The number of updates is set to 8, 16, 32, or 64. 
Learning rates of 1.0, 0.1, 0.01, and 0.001 are compared based on average accuracy across 100 fine-tuning runs with 100 training tasks, and 0.01 is chosen because it achieves perfect 1.0 accuracy with 32- and 64-step updates and the best accuracies with 8- and 16-step updates (0.970 and 0.999).
After each of the 8192 fine-tuning runs, the diagonal elements of the Fisher information matrix are computed and averaged across the components of each network filter, yielding the task embedding vector, as prescribed in ~\citep{achille2019task2vec}.
\item Embedding test tasks: Using 1024 test tasks of FC60, we compute the task embedding vectors with the same pipeline and hyperparameters as in embedding the training tasks. 
After that, the cosine similarity values between the 8192 x 1024 pairs of training and test task embeddings are computed. 
\end{itemize}

Table \ref{tab: label_overlap} shows the computation time each algorithm spends for scoring the similarities of the 8192 x 1024 pairs of training task and test task embeddings. This computation time corresponds to the one spent in user environments in future real-world applications.
In each case of MAML + TLXML or Protonet + TLXML, it refers to the time spent for computing the influences of the 8192 training tasks on each of 1024 test losses after $I^{\rm meta}$ is determined, that is, computing \mEqref{eq:influence_ml_gen_perf} with a given $I^{\rm meta}$.
In each case of Task2Vec, it refers to the time spent on 1024 fine-tuning runs with 1024 test tasks, computing their embedding vectors, and computing cosine similarities with the 8192 training tasks. 
The number in the brackets indicates the number of iterations over which the network head is updated in each fine-tuning run. 
In the table, we observe that TLXML outperforms Task2Vec with 8, 16, 32, and 64 fine-tuning steps in terms of computation time.
Moreover, as we will see below, among the four Task2Vec settings, only the 64-step setting is on par with TLXML in terms of the consistency with semantic similarity. TLXML is 155 sec / 33 sec$\sim 5$ or 155 sec / 56 sec$\sim 3$ times faster than Task2Vec with that setting.

In addition, we consider classifying the labels in a test task and defining the training subpopulation for each classified label group.  
Table
\ref{tab: label_overlap_recall_maml}, and
\ref{tab: label_overlap_recall_protonet}
show the results from counting the proper tests when the training subpopulation is defined based on the number of shared superclasses with the test labels classified by their recall values achieved in the tests. 
Although the classification reduces the number of statistics per row, the Tables again demonstrate that TLXML identifies subpopulations in the training taskset that share semantic properties with test tasks. 
Furthermore, we observe a tendency for labels with low recall to define subpopulations of comparatively low significance, implying a correlation between recall and influence scores (i.e., when a label has high recall, training tasks with labels in the same superclass are scored highly).

\subsection{One-Step Update Using TLXML}
\label{sup: one_step_update}
\begin{table*}[th]
\centering
\caption{Test accuracies after single TLXML-guided updates to MAML-trained models with blocked and enhanced tasks.
$\pm$ means the average and standard deviation across 5 runs of MAML training. 
Bold values outperform the MAML baseline (shown in the brackets) with a unpaired Welch two-sample \(t\)-test, \(p<0.05\). $\dagger$ means the average is below the baseline value.  
}

\begin{tabular}{lccccc}
\toprule
\textbf{train$\rightarrow$test} &
\textbf{\# tasks}  &
{} &
{} &
{} &
\textbf{leave-out} \\
\midrule
\shortstack[l]{FC60$\rightarrow$FC60}
   & 128
   & {}
   & {}
   & {}
   & 0.662$\pm$0.006$^\dagger$\\
($0.663\pm0.004$) 
   & 256
   & {}
   & {}
   & {}
   & 0.660$\pm$0.007$^\dagger$ \\
   & 512
   & {}
   & {}
   & {}
   & 0.663$\pm$0.008 \\
   & 1024
   & {}
   & {}
   & {}
   & 0.661$\pm$0.006$^\dagger$ \\
   & 2048
   & {}
   & {}
   & {}
   & 0.666$\pm$0.010 \\
\midrule
\textbf{train$\rightarrow$test} &
\textbf{\# tasks} &
\textbf{block} & \textbf{block} & \textbf{block} & \textbf{block} \\
{} & {} &
$\boldsymbol{\xi=-8}$ &
$\boldsymbol{\xi=-4}$ &
$\boldsymbol{\xi=-2}$ &
$\boldsymbol{\xi=-1}$ \\
\midrule
\shortstack[l]{FC60$\rightarrow$FC60}
   & 128
   & 0.670$\pm$0.005
   & 0.667$\pm$0.004
   & 0.665$\pm$0.004
   & 0.664$\pm$0.004\\
($0.663\pm0.004$) 
   & 256
   & \textbf{0.672$\pm$0.005}
   & 0.670$\pm$0.004
   & 0.666$\pm$0.004
   & 0.665$\pm$0.004\\
   & 512
   & \textbf{0.675$\pm$0.006}
   & \textbf{0.672$\pm$0.005}
   & 0.669$\pm$0.004
   & 0.666$\pm$0.004\\
   & 1024
   & 0.673$\pm$0.008
   & \textbf{0.676$\pm$0.005}
   & \textbf{0.671$\pm$0.004}
   & 0.668$\pm$0.004\\
   & 2048
   & 0.659$\pm$0.009
   & \textbf{0.674$\pm$0.007}
   & \textbf{0.672$\pm$0.004}
   & 0.669$\pm$0.004\\
\cmidrule(lr){1-6}
\shortstack[l]{FC60$\rightarrow$FC100}
   & 128
   & \textbf{0.439$\pm$0.007}
   & 0.436$\pm$0.005
   & 0.433$\pm$0.005
   & 0.431$\pm$0.005\\
($0.429\pm0.005$)
   & 256
   & \textbf{0.443$\pm$0.008}
   & \textbf{0.438$\pm$0.006}
   & 0.435$\pm$0.006
   & 0.432$\pm$0.005\\
   & 512
   & \textbf{0.448$\pm$0.011}
   & \textbf{0.443$\pm$0.007}
   & \textbf{0.438$\pm$0.005}
   & 0.434$\pm$0.005\\
   & 1024
   & \textbf{0.452$\pm$0.011}
   & \textbf{0.448$\pm$0.009}
   & \textbf{0.441$\pm$0.007}
   & 0.437$\pm$0.006\\
   & 2048
   & 0.447$\pm$0.013
   & \textbf{0.449$\pm$0.011}
   & \textbf{0.444$\pm$0.008}
   & 0.439$\pm$0.007\\
\midrule
\textbf{train$\rightarrow$test} &
\textbf{\# tasks} &
\textbf{enhance} & \textbf{enhance} & \textbf{enhance} & \textbf{enhance} \\
{} & {} &
$\boldsymbol{\xi=1}$ &
$\boldsymbol{\xi=2}$ &
$\boldsymbol{\xi=4}$ &
$\boldsymbol{\xi=8}$ \\
\midrule
\shortstack[l]{FC60$\rightarrow$FC60}
   & 128
   & 0.664$\pm$0.004
   & 0.666$\pm$0.005
   & 0.668$\pm$0.005
   & \textbf{0.672$\pm$0.005}\\
($0.663\pm0.004$)
   & 256
   & 0.666$\pm$0.005
   & 0.668$\pm$0.005
   & \textbf{0.672$\pm$0.005}
   & \textbf{0.676$\pm$0.005}\\
   & 512
   & 0.668$\pm$0.005
   & \textbf{0.672$\pm$0.005}
   & \textbf{0.677$\pm$0.006}
   & \textbf{0.677$\pm$0.007}\\
   & 1024
   & \textbf{0.670$\pm$0.005}
   & \textbf{0.676$\pm$0.005}
   & \textbf{0.680$\pm$0.006}
   & 0.662$\pm$0.010$^\dagger$\\
   & 2048
   & \textbf{0.675$\pm$0.005}
   & \textbf{0.680$\pm$0.006}
   & \textbf{0.676$\pm$0.008}
   & \textbf{0.595$\pm$0.026}$^\dagger$\\
\cmidrule(lr){1-6}
\shortstack[l]{FC60$\rightarrow$FC100}
   & 128
   & 0.432$\pm$0.005
   & 0.433$\pm$0.004
   & 0.436$\pm$0.005
   & 0.437$\pm$0.007\\
($0.429\pm0.005$)
   & 256
   & 0.434$\pm$0.005
   & 0.436$\pm$0.005
   & 0.438$\pm$0.007
   & 0.436$\pm$0.012\\
   & 512
   & 0.435$\pm$0.005
   & 0.438$\pm$0.007
   & 0.439$\pm$0.010
   & 0.430$\pm$0.012\\
   & 1024
   & \textbf{0.438$\pm$0.005}
   & \textbf{0.441$\pm$0.008}
   & 0.437$\pm$0.011
   & \textbf{0.413$\pm$0.008}$^\dagger$\\
   & 2048
   & \textbf{0.440$\pm$0.007}
   & \textbf{0.442$\pm$0.009}
   & 0.429$\pm$0.010
   & \textbf{0.390$\pm$0.020}$^\dagger$\\
\bottomrule
\end{tabular}
\label{tab: onestep_maml_all}
\end{table*}
We examine the effect of the one-step update \mEqref{eq: one_step_update} from the convergence point of meta-learning.
We train a 3-conv$\times$16-filter network(6469 parameters) with MAML, using 8192 training tasks from the FC60 dataset and 40,000 meta-batches. 
We conduct 5 training runs with seeds 0, 1, 2, 3, and 42.
Each training task is scored using the influence function for the average test loss across 1024 test tasks.
For each setting of the number of blocked or enhanced tasks, those tasks are determined by selecting the worst or best training tasks from the scoring results.

Table \ref{tab: onestep_maml_all} shows the effects of both blocking and enhancing training tasks, and the impact on the test accuracies in two cases: test datasets with and without shared superclasses (FC60 test taskset and FC100 test taskset, respectively). 
For each combination of blocked/enhanced and the two test tasksets, we observe cases of improved test accuracy.
The improvement is slight when the absolute value of the shift parameter $\xi$ is small, and there are few blocked/enhanced tasks, as that is close to doing nothing.
The improvement is also slight in the region of a large absolute value of the shift parameter $\xi$ and a large number of blocked/enhanced tasks, which is considered to be caused by the deterioration of the linear approximation, an underlying assumption in using the influence function $I^{\rm{meta}}$.

The table also shows the results of leave-out retraining. For each set of blocked tasks, the network was trained from scratch with those tasks removed from the dataset. We observe that the leave-out trainings fail to yield a statistically significant gain.

\clearpage

\subsection{Runtime and peak memory usage} \label{sup: runtime}

Using the synthetic dataset described in \ref{sup: dataset}, we measure the runtimes and peak values of system and GPU memory usage for calculating $I^{\rm meta}$ for MAML as we increase the network width and depth. For each network structure, we conduct measurements for two cases: the exact Hessian (\mEqref{eq: hessian-ml-gen}) and the Gauss-Newton approximation of it(\mEqref{eq: OPA}).
When neither results in an out-of-memory (OOM) error, we calculate the Pearson correlation between the influence scores of the training tasks on each test loss for the two. 
In both cases, we employ the pseudo-inverse of the Hessians for defining $I^{\rm meta}$ (\mEqref{eq:influence_ml_gen_param_mod}) and the following settings:
\begin{itemize}
\item Problem setting: 3-way-5-shot problem
\item  Training dataset size: 1024 tasks
\item  Network type: fully connected network
\item  Meta batch size: 128
\item Train iterations: meta-parameters are updated with 5000 meta-batches
\end{itemize}

As mentioned in \ref{sup: implementation}, the rank of the Hessian corresponds to the number of constraints imposed by the training taskset. 
Using a network with 4-4 hidden layers, we searched for the optimal rank value regarding the resulting self-rank and estimated it to be 22.
The results of the measurements with that estimated rank are listed in Table \ref{tab: width_scaling} and Table \ref{tab: depth_scaling}.
We observe that the runtimes and GPU memory costs with the exact Hessian become intractable when the network size is large, as expected in \mref{hessian_approximation} and \ref{sup: explicit_forms}. 
In contrast, those with the approximated Hessian(GN) remain tractable. We should also note that the influence scores with the approximated Hessian (GN) show good correlation with the exact Hessian (Table \ref{tab: corr_width_scaling} and Table \ref{tab: corr_depth_scaling}).  

\begin{table}[t]
		
\centering
\caption{Runtime and peak memory usage (width scaling)}
\label{tab: width_scaling}
\begin{tabular}{cccccccc}
	\toprule
	Widths & Params & \multicolumn{3}{l}{Exact} & \multicolumn{3}{l}{Gauss-Newton} \\
	{}&{}& runtime & sys-mem & gpu-mem &runtime & sys-mem & gpu-mem \\
	\midrule	
	$[4, 4]$  &63   &151.6 s & 1006 MiB&  18 MiB&  86.8 s& 1004 MiB&   17 MiB \\
	$[8, 8]$  &155  &330.9 s & 1048 MiB&  23 MiB&  86.5 s& 1006 MiB&   17 MiB \\
	$[16, 16]$&435  &861.8 s & 1180 MiB&  54 MiB&  86.5 s& 1012 MiB&   17 MiB \\
	$[32, 32]$&1379 &2682.0 s & 1663 MiB& 287 MiB&  86.1 s& 1012 MiB&   17 MiB \\
	$[64, 64]$&4803 &10063.9 s & 4451 MiB&2405 MiB&  90.4 s& 1029 MiB&   18 MiB \\
	$[96, 96]$&10275&- & - &  OOM              &  96.3 s& 1150 MiB&   18 MiB \\
	\bottomrule
\end{tabular}

\centering
\caption{Runtime and peak memory usage (depth scaling)}
\label{tab: depth_scaling}
\begin{tabular}{cccccccc}
	\toprule
	Widths & Params & \multicolumn{3}{l}{Exact} & \multicolumn{3}{l}{Gauss-Newton} \\
	{}&{}& runtime & sys-mem & gpu-mem &runtime & sys-mem & gpu-mem \\
	\midrule	
	$[32, 32]$     &1379&  2682.0 s & 1663 MiB& 287 MiB&  86.1 s& 1012 MiB&  17 MiB \\
	$[32, 32, 32]$ &2499&  7438.1 s & 2863 MiB& 796 MiB& 116.0 s& 1018 MiB&  17 MiB \\
	$[32]\times 5$ &4739&  23695.8 s &  7553 MiB&2577 MiB& 179.7 s& 1100 MiB&  18 MiB \\
	$[32]\times 9$ &9219&  - &  - & OOM & 298.3 s& 1192 MiB&  18 MiB \\
	\bottomrule
\end{tabular}

\centering
\caption{Influence score correlation (exact vs Gauss-Newton, width scaling)}
\label{tab: corr_width_scaling}
\begin{tabular}{cccccc}
	\toprule
	Widths & Params & \multicolumn{2}{l}{Accuracy} & Correlation \\
	{}&{}&{train}&{test}& \\
	\midrule	
	$[4, 4]$  &63   &0.975&0.974& 0.652$\pm$0.173 \\
	$[8, 8]$  &155  &0.985&0.980& 0.805$\pm$0.088 \\
	$[16, 16]$&435  &0.989&0.983& 0.737$\pm$0.131 \\
	$[32, 32]$&1379 &0.992&0.985& 0.780$\pm$0.150 \\
	$[64, 64]$&4803 &0.996&0.987& 0.736$\pm$0.142 \\
	$[96, 96]$&10275&0.995&0.986& -    \\
	\bottomrule
\end{tabular}
\centering
\caption{Influence score correlation (exact vs Gauss-Newton, depth scaling)}
\label{tab: corr_depth_scaling}
\begin{tabular}{cccccc}
	\toprule
	Widths & Params & \multicolumn{2}{l}{Accuracy} & Correlation \\
	{}&{}&{train}&{test}& \\
	\midrule	
	$[32, 32]$     &1379&0.992&0.985& 0.780$\pm$0.150\\
	$[32, 32, 32]$ &2499&0.993&0.985& 0.734$\pm$0.117\\
	$[32]\times 5$ &4739&0.996&0.985& 0.803$\pm$0.085\\
	$[32]\times 9$ &9219&0.998&0.985& -          \\
	\bottomrule
\end{tabular}
\end{table}

\subsection{Computational Resources}
The experiments in this paper are conducted with multiple computing environments. In a typical environment, the machine is equipped with an Intel Core i7-7567U CPU running at 3.50 GHz, 32GB RAM, and an NVIDIA GeForce RTX 2070 GPU, and runs Ubuntu 24.04 LTS.
The machine used for measuring runtimes and memory usage is equipped with an Intel Core i5-8400 CPU running at 2.80GHz,  an NVIDIA GeForce GTX 1080 GPU (8GB RAM), and Ubuntu 20.04 LTS.

\ResetLabel

\end{document}